\documentclass[preprint,12pt,authoryear]{elsarticle}
\usepackage{graphicx}
\usepackage{graphics}
\usepackage{subfigure}
\usepackage{epstopdf}
\usepackage{csquotes}
\newcommand{\ket}[1]{|#1\rangle}
\newcommand{\bra}[1]{\langle#1|}

\usepackage[toc,title,page]{appendix}
\usepackage{mathrsfs}
\usepackage{mathtools}
\usepackage{subfigure}
\usepackage{amsmath}
\DeclareMathOperator{\trace}{tr}

\let\oldcite\cite
\renewcommand*\cite[1]{[\oldcite{#1}]}
\usepackage{times}
\usepackage{soul}
\usepackage{verbatim}
\usepackage{algcompatible}
\usepackage{epsfig}
\usepackage{amssymb}
\usepackage{makeidx}
\journal{Neural Networks}

\begin{document}

\begin{frontmatter}

\title{A hybrid quantum-classical neural network with deep residual learning}

\author[label1,label2]{Yanying Liang\corref{cor1}}
\author[label2]{Wei Peng}
\author[label1,label3]{Zhu-Jun Zheng}
\author[label2]{Olli Silv\'{e}n}
\author[label2]{Guoying Zhao}

\address[label1]{School of Mathematics, South China University of Technology, Guangzhou 510641, China}
\address[label2]{Center for Machine Vision and Signal Analysis, University of Oulu, Oulu 90570, Finland}
\address[label3]{Laboratory of Quantum Science and Engineering, South China University of Technology, Guangzhou 510641, China}

\cortext[cor1]{Corresponding author: yanying.liang@oulu.fi}

\begin{abstract}
\small{Inspired by the success of classical neural networks, there has been tremendous effort to develop classical effective neural networks into quantum concept. In this paper, a novel hybrid quantum-classical neural network with deep residual learning (Res-HQCNN) is proposed. 
We firstly analysis how to connect residual block structure with a quantum neural network, and give the corresponding training algorithm. At the same time, the advantages and disadvantages of transforming deep residual learning into quantum concept are provided. As a result, the model can be trained in an end-to-end fashion, analogue to the backpropagation in classical neural networks.
 To explore the effectiveness of Res-HQCNN , we perform extensive experiments for quantum data with or without noisy on classical computer. The experimental results show the Res-HQCNN performs better to learn an unknown unitary transformation and has stronger robustness for noisy data, when compared to state of the arts. Moreover, the possible methods of combining residual learning with quantum neural networks are also discussed.}
\end{abstract}

\begin{keyword}
\small{quantum computing \sep quantum neural networks \sep deep residual learning}


\end{keyword}

\end{frontmatter}


\section{Introduction}

Artificial neural networks (ANNs) stand for one of the most prosperous computational paradigms \cite{nielsen2015neural}.  Early neural networks can be traced back to the McCulloch-Pitts neurons in 1943 \cite{mcculloch1943logical}. Over the past few decades, taking advantage of a number of technical factors, such as new and scalable software platforms \cite{bergstra2010theano,jia2014caffe,maclaurin2015autograd,paszke2017automatic,paszke2019pytorch} and powerful special-purpose computational hardware \cite{chetlur2014cudnn,jouppi2017datacenter}, the development of machine learning techniques based on neural networks makes progress and breakthroughs. Currently, neural networks achieve significant success and have wide applications in various machine learning fields like pattern recognition, video analysis, medical diagnosis, and robot control \cite{bishop1995neural,nishani2017computer,amato2013artificial,mitchell1993explanation}. One notable reason is the increasingly passion for exploring new neural architectures, including manual ways by expert knowledge and automatic ways by auto machine learning~\cite{he2016deep,pham2018efficient,peng2019learning}. Particularly, in 2016, deep residual networks (ResNets) are proposed with extremely deep architectures showing excellent accuracy and nice convergence behaviors \cite{he2016deep,he2016identity}. Their result won the 1st place on the ImageNet Large Scale Visual Recognition Challenge 2015 classification task for ImageNet classification and ResNet (and its variants) achieves revolutionary success in many research and industry applications.

In parallel with the development of ANNs, quantum neural networks (QNNs) appear with the potential to evade the limitation of the classical computation due to the use of quantum computing. Compared with classical computing, quantum computing has possible advantages with the properties of quantum mechanics, such as quantum entanglement, quantum superposition and massive parallelism \cite{biamonte2017quantum,ronnow2014defining}. Earlier quantum neural networks~\cite{purushothaman1997quantum,ezhov2000quantum} can even trace back to two decades ago. As the current quantum technologies and devices developing, there is a huge passion to combine quantum computing with ANNs for quantum data with or without noisy \cite{schuld2014quest,dunjko2018machine,sasaki2002quantum,dunjko2016quantum,alvarez2017supervised,verdon2018universal,sentis2019unsupervised,zhao2019building,li2020quantum,bisio2010optimal,sedlak2019optimal,beer2020training,2021two,2021three,2021four}.

\begin{figure*}[htbp]
\centering
\subfigure[clean data]{
\includegraphics[width=6cm]{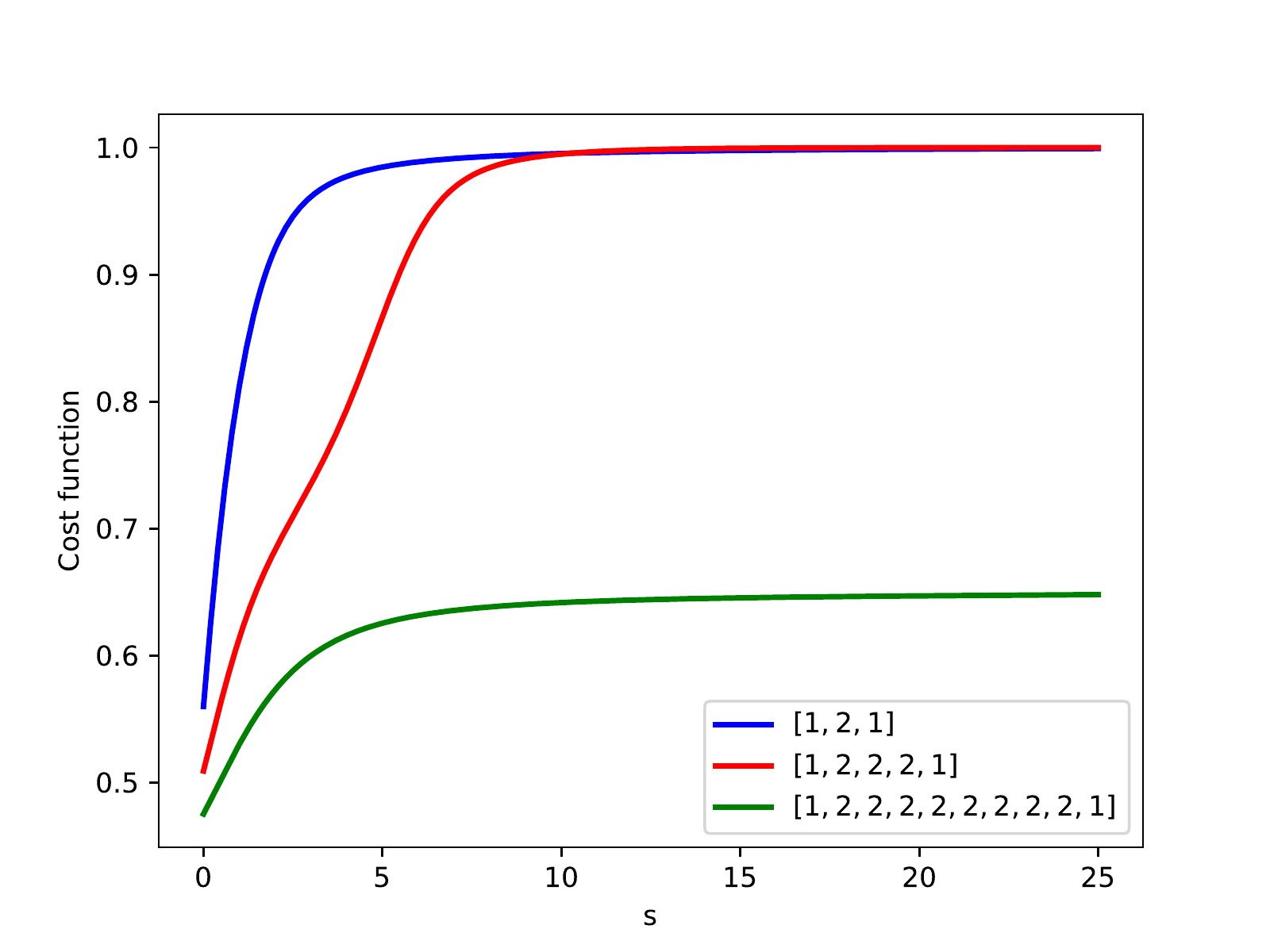}
}
\subfigure[noisy data]{
\includegraphics[width=6cm]{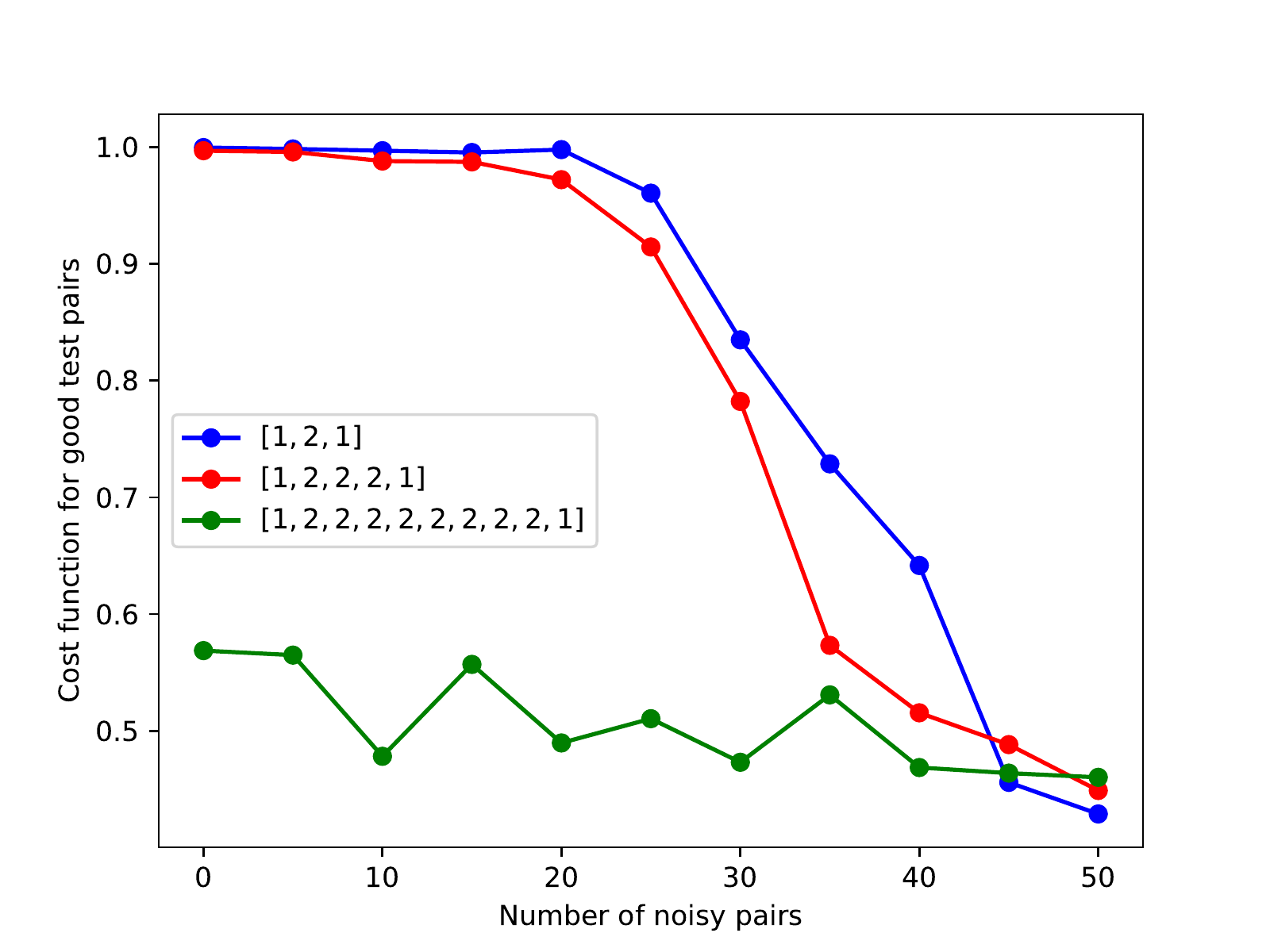}
}
\caption{ \textbf{Numerical results of QNNs in \cite{beer2020training}}.}
\label{Figure1}
\end{figure*}

Among these works, \cite{beer2020training} caught our attention. On one hand, it provides the readers with accessible code. On the other hand, the paper presents a truly quantum neurons forming a quantum feedforward neural network. It has
 remarkable ability to learn an unknown unitary and striking robustness to noisy training data. This work is important for noisy intermediate-scale quantum devices due to the reduction in the number of coherent qubits. We are interested in it. The cost function in this work is chosen as the fidelity between a pure quantum state and an arbitrary quantum state. Nevertheless, 
 as the number of network layers deepens, we find that the rate of convergence of the cost function decreases, and the value of convergence even can not reach the maximum for clean data, see Fig. \ref{Figure1}(a). As for noisy data, Fig. \ref{Figure1}(b) shows that the robustness for noisy data gets weaker with the network deeper and deeper.
 So we wonder whether the performance of the cost function for both clean and noisy data can be improved or not. Here we use a 1-dimensional list of natural numbers to represent the number of perceptrons in the corresponding layer.
 
Inspired by the efficiency of deep residual learning, we try to design a novel quantum-classical neural network with deep residual learning (Res-HQCNN) to achieve our goal. This idea is novel, and as far as we know, no work has been attempted up to now. We wish to explore how to put residual scheme into the QNNs in \cite{beer2020training} efficiently. It is not trivial. First, the QNNs in \cite{beer2020training} is a closed quantum system. The dynamics of a closed quantum system are described by a unitary transform. The input and output matrices in the QNNs of Beer \emph{et al.} have unit trace, which is an important constraint for a density matrix. The residual scheme may increase the trace of input and output matrix, which will cause that the implementation is not possible for quantum computer. Second, our goal is to improve the performance of the cost function for both clean and noisy quantum data. We hope the experiment can not only be carried out on quantum computer, but also have improved performance. Third, connecting residual scheme into a feed-forward QNNs will change the training algorithm, especially the process of updating unitary perceptrons. The number of the residual block structure, the number of network layers, and the choice of skipping layers or not will all influence the procedure of updating parameters. Since the updating parameters matrix is calculated from the definition of derivative function, different network structure has different updating parameters matrix. The more diverse the network structure, the more complex the corresponding updating parameters matrix. So this exploration is challenging but interesting. We hope our paper can be an useful reference in this research area. Contributions stemming from this paper include:
 \begin{itemize}
     \item Design a new residual learning structure based on the QNNs in \cite{beer2020training}.
     \item Give the model of Res-HQCNN and calculate the new training algorithm. Based on the training algorithm, we present an analysis from the perspective of propagating information feedforward and backward.
     \item Explore different ways of connection between the new residual scheme and QNNs, such as identity shortcut connection skipping one layer.
     \item Present improved performance of Res-HQCNN on both clean and noisy quantum data over the former QNNs at the cost of implementing only on classical computer.
     \item Discuss another method to design residual block structure into quantum neural networks so that the implementation can be carried out on quantum computer.
 \end{itemize}
 
 The remainder of this paper is organized as following. Section \ref{2} reviews the related contributions about quantum neural networks and residual scheme. In Section \ref{3}, we briefly introduce the basic concepts of quantum qubits and the operators we mainly use in the paper, as well as the mechanism of deep residual learning. Section \ref{4} gives the model of quantum neural network with deep residual learning, including its architecture and training algorithm on classical computer. To validate the improved performance, Section \ref{5} provides the experimental simulations and corresponding analysis. Finally conclusion and discussion in Section \ref{6} are given.

\section{Related work}\label{2}

\subsection{Quantum Neural Networks}
Quantum neural networks have strong potential to be superior to the classical neural network after combining neural computing with the mechanics in quantum computing. Quantum data is in the form of quantum states. Just as a classical bit has a state 0 or 1, a qubit also has a state. Two possible state for a qubit are the states $\ket{0}$ and $\ket{1}$. The examples can be the two different polarizations of a photon and two states of an electron orbiting a single atom. Quantum neural networks can also process real-world data \cite{review1image,review2image}. In this paper, we mainly focus on quantum neural networks with quantum data.
Specifically, these research achievements mainly include the following aspects:
solving central tasks in quantum learning \cite{sasaki2002quantum,bisio2010optimal,beer2020training};
enhancing the problem of machine learning  \cite{dunjko2016quantum,alvarez2017supervised,purushothaman1997quantum};
and efficient classification of quantum data \cite{sentis2019unsupervised,zhao2019building,li2020quantum}.
Among them, those papers about quantum learning for an unknown unitary transformation impress us a lot. In detail,
Bisio and Chiribella \cite{bisio2010optimal} addressed this task and found optimal strategy to store and retrieve an unknown unitary transformation on a quantum memory. Soon after, Sedl{\'a}k \emph{et al.} \cite{sedlak2019optimal} designed an optimal protocol of unitary channels, which generalizes the results in \cite{bisio2010optimal}. Moreover, Beer \emph{et al.} \cite{beer2020training} proposed a quantum neural network with remarkable generalisation behaviour for the task of learning an unknown unitary quantum transformation.
We hope the ability of learning unknown unitary transformation can be improved due to the new idea we propose.

\subsection{Residual Scheme in Neural Networks}
Proposed in 2012, AlexNet~\cite{krizhevsky2017imagenet} became one of the most famous neural network architecture in deep learning era. This was treated as the first time that deep neural network was more successful than traditional, hand-crafted feature learning on the ImageNet~\cite{deng2009imagenet}.  Since then, researchers pay many efforts to make the network deeper, as deeper architecture could potentially extract more important semantic information. But deeper networks are more difficult to train, due to the notorious vanishing/exploding gradient problem. Residual scheme in ResNet~\cite{he2016deep} is one of the most successful strategy of improving current neural networks. ResNet makes it possible to train up to hundreds or even thousands of layers and still achieves compelling performance. Basically, this scheme reformulates the layers as learning residual functions with reference to the layer inputs, instead of learning unreferenced functions. Based on this, a residual neural network builds on constructs known from pyramidal cells in the cerebral cortex, utilizing skip connections, or shortcuts to jump over some layers. This simple but efficient strategy largely improves the performance of current neural architectures in many fields. For instance, in image classification tasks, many variants of ResNets~\cite{he2016deep,xie2017aggregated,chen2018neural,dong2020towards,korpi2020deeprx} are proposed and get the state-of-the-art performance. This scheme is even introduced into graph convolutional networks. For instance, works from~\cite{yan2018spatial,peng2019learning,peng2020mix} also endow graph convolutional network with the residual connections, capturing a better representations for skeleton graphs.
Nevertheless, as far as we know, there is no work introducing the residual scheme in the field of quantum neural networks. In this paper, we will make the first attempt to do this and present an efficient Res-HQCNN which can be trained with an end-to-end fashion.

\section{ Preliminaries}\label{3}
\subsection{ Qubits and quantum operators}
Analogous to the role bit is the smallest unit of classical computing, qubit is the smallest unit in quantum computing. The notation $\ket{\cdot}$ is called a ket which is used to indicate that the object is a column vector. The complex conjugate transpose of $\ket{\cdot}$ is $\bra{\cdot}$, which is called a bra \cite{nielsen2002quantum}.
A two-level quantum system in a two-dimensional Hilbert space $\mathbb{C}^2$
with basis $\{\ket{0}, \ket{1}\}$ is a single qubit, which can be written from the superposition principles:
$$\ket{\psi}=\alpha\ket{0}+\beta\ket{1}$$
$$|\alpha|^2+|\beta|^2=1, \alpha, \beta \in \mathbb{C}.$$

The quantum operators used in this paper mainly include tensor product operator, reduced density operator and partial trace.
Tensor product is a way to extend the dimension of vector spaces through putting vector space together \cite{nielsen2002quantum}. The symbol for tensor product is denoted by $\otimes$. Assume $A$ is an $m$ by $n$ matrix, $B$ is a $p$ by $q$ matrix, then $A\otimes B$ is an $m\cdot p$ by $n\cdot q$ matrix:
$$A\otimes B=\begin{bmatrix}
A_{11}B & A_{12}B &\cdots  & A_{1n}B\\
A_{21}B &A_{22}B  & \cdots & A_{2n}B\\
\vdots & \vdots & \vdots & \vdots\\
 A_{m1}B& A_{m2}B &\cdots  & A_{mn}B
\end{bmatrix}.$$

Another important operator used in this paper is the reduced density operator \cite{nielsen2002quantum}. It is often used to get the desired subsystems of a composite quantum system. Assume $A$ and $B$ are two physical systems. The state in $A\otimes B$ is described by density matrix $\rho_{AB}$. The reduced density operator for $A$ is given by
$$\rho_{A}=\trace_{B}(\rho_{AB}),$$
where $\trace_{B}$ is a map of operators called partial trace over $B$. Here partial trace is defined as
$$\trace_{B}(\ket{a_1}\bra{a_{2}}\otimes \ket{b_{1}}\bra{b_{2}})=\ket{a_1}\bra{a_{2}}\trace(\ket{b_{1}}\bra{b_{2}}),$$
where $\ket{a_1}$ and $\bra{a_{2}}$ are any vectors in $A$, $\ket{b_1}$ and $\bra{b_{2}}$ are any vectors in $B$.
For example, if $A$ and $B$ are both two-dimensional complex vector space $\mathbb{C}^2$, then
$$\rho_{A}=\trace_{B}(\ket{0}_A\bra{0}\otimes \ket{0}_B\bra{0})=\ket{0}_A\bra{0} \trace(\ket{0}_B\bra{0})=\begin{bmatrix}
1 &0 \\
 0& 0
\end{bmatrix},$$
$$\rho_{B}=\trace_{A}(\ket{0}_A\bra{0}\otimes \ket{1}_B\bra{1})=\ket{1}_B\bra{1} \trace(\ket{0}_A\bra{0})=\begin{bmatrix}
0 &0 \\
 0& 1
\end{bmatrix}.$$
Here $\ket{0}_A\bra{0}$ means $\ket{0}\bra{0}$ in physical system $A$.
\subsection{Deep residual learning}

\begin{figure}[ht]
\centering
\includegraphics[angle=0,width=0.7\linewidth]{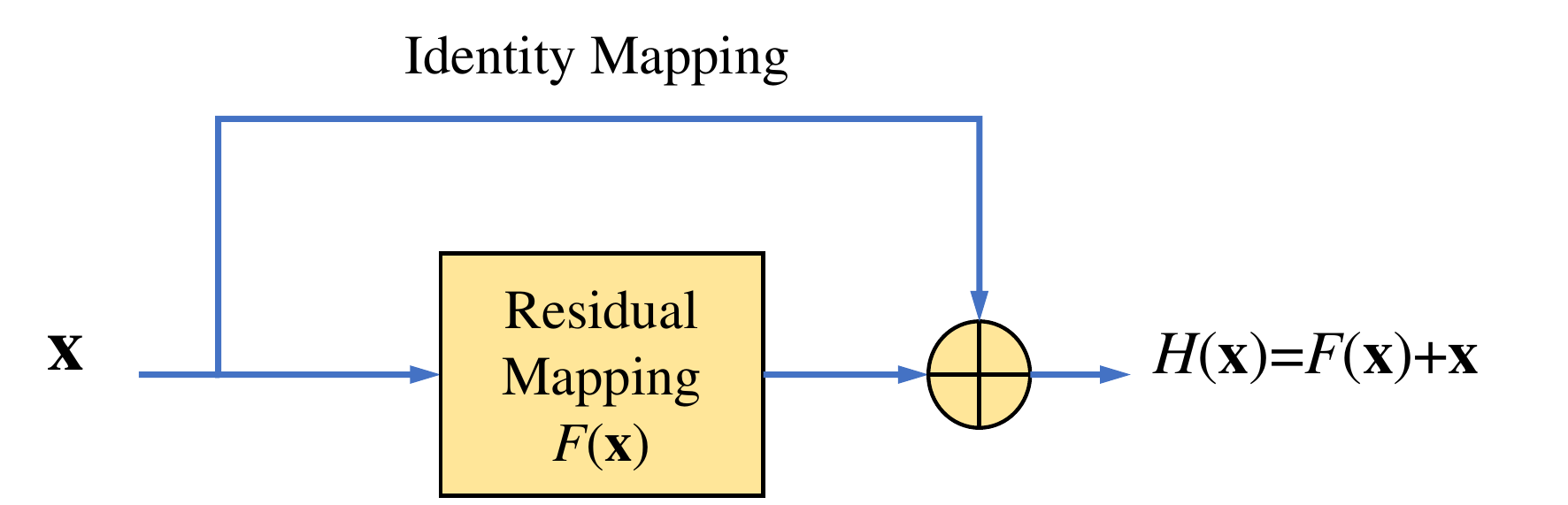}
\caption{\textbf{Residual block structure.}}
\label{fig2}
\end{figure}

Training a deep neural network can be computationally costly. He \emph{et al.} \cite{he2016deep} proposed neural networks with deep residual learning framework, which is easier to optimize and obtain high accuracy from increased depth. In a residual block structure, there is a shortcut pathway connecting the input and output of a block structure. Specifically, residual learning chooses to fit the residual mapping \textit{F}(\textsc{x}):=\textit{H}(\textsc{x})-\textsc{x}, rather than approximating the desired underlying mapping \textit{H}(\textsc{x}) directly. The final mapping of a residual block structure is \textit{F}(\textsc{x})+\textsc{x}, which is equivalent to \textit{H}(\textsc{x}), see Fig. \ref{fig2}. 

The degradation problem tells us that the deeper network has higher training error and test error. But compared with shallower counterpart, the performance of deeper network should not be worse. This suggests that the network might have difficulties in approximating identity mapping. So optimizing the residual mapping \textit{F}(\textsc{x}) towards zero is relatively easier than approximating \textit{H}(\textsc{x}) into identity mapping.

For ANNs, deep residual learning has the advantage of solving the problem of vanishing/exploding gradients and the degradation problem. This paper is mainly based on the QNNs in \cite{beer2020training}, which is absence of a barren plateau in the cost function landscape. So we expect the QNNs with residual learning can get improved performance of the cost function both for clean and noisy data. We hope that the deeper the Res-HQCNN, the more effective it will be.

\section{ The model of Res-HQCNN}\label{4}

In this section, we define the architecture of Res-HQCNN based on the QNNs in \cite{beer2020training}. According to the mechanism of the defined Res-HQCNN, we explain its training algorithm in different cases.

\subsection{ The architecture of Res-HQCNN}
We firstly define a residual block structure in Res-HQCNN. Then the architecture of Res-HQCNN with multiple layers is presented. For better understanding the mechanism, we show an example of Res-HQCNN with one hidden layer. Finally, we give an analysis about the differences between former QNNs and Res-HQCNN.

Based on the residual block structure in Fig. \ref{fig2}, a new residual block structure in Res-HQCNN is defined as following.
For convenience, we put forward a few assumptions and notations at the beginning.
\begin{figure}[ht]
\centering
\includegraphics[angle=0,width=0.8\linewidth]{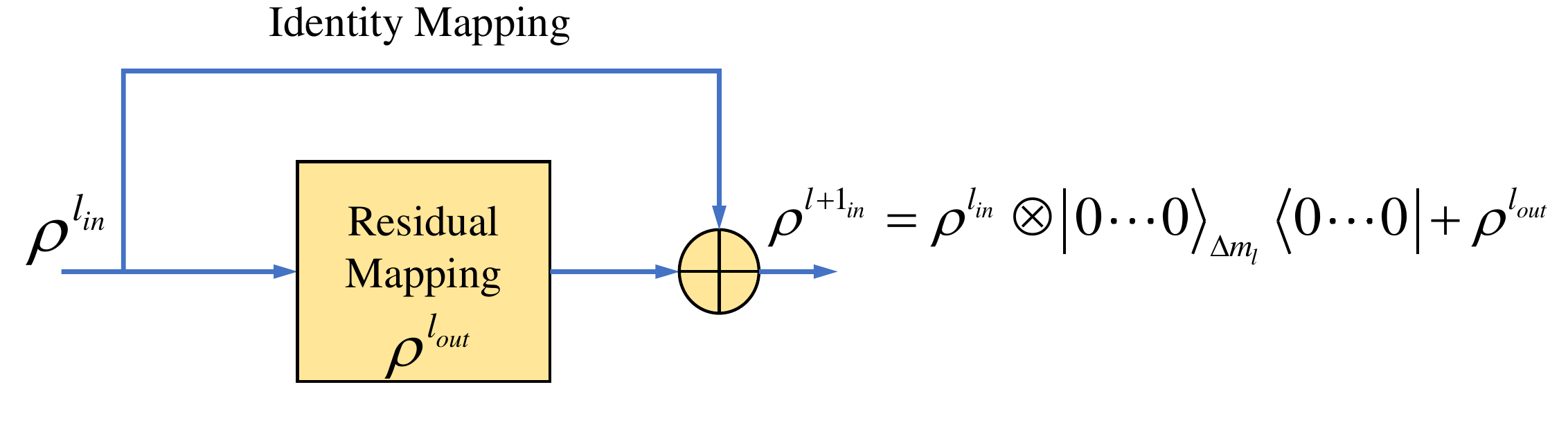}
\caption{\textbf{Residual block structure in Res-HQCNN.} Here $\Delta m_l=m_l-m_{l-1}.$ }
\label{fig3}
\end{figure}
 Res-HQCNN has $L$ hidden layers. The perceptron nodes of each layer represent single qubits. Denote $m_l$ as the number of nodes in each layer $l$  and we assume $m_{l-1} \leq m_l$ for $l=1,2,\cdots,L$. Here $l=0$ represents the input layer and $l=L+1$ corresponds to output layer. Denote $\rho^{l_{in}}$ and $\rho^{l_{out}}$ represent the input and output state of layer $l$ in Res-HQCNN. Then the residual block structure can be designed in Fig. \ref{fig3}.

 Mathematically, we set the input mapping and final mapping of the residual block structure in Res-HQCNN as $\rho^{l_{in}}$ and $\rho^{{l+1}_{in}}$, respectively. The residual mapping is chosen as $\rho^{l_{out}}$. It is important to note that the output of the former layer is not the exact input of the next layer. As shown in Fig. \ref{fig3}, the new input of layer $l+1$ is the addition of the output state and the input state in layer $l$. Here the additive operation corresponds to the matrix element-wise addition. Due to $m_{l-1} \leq m_l$, We apply tensor product to $\rho^{l_{in}}$ and $\ket{0\cdots0}_{\Delta m_l}\bra{0\cdots0}$ to keep the same dimension.
 
\begin{figure}[ht]
\centering
\includegraphics[angle=0,width=0.8\linewidth]{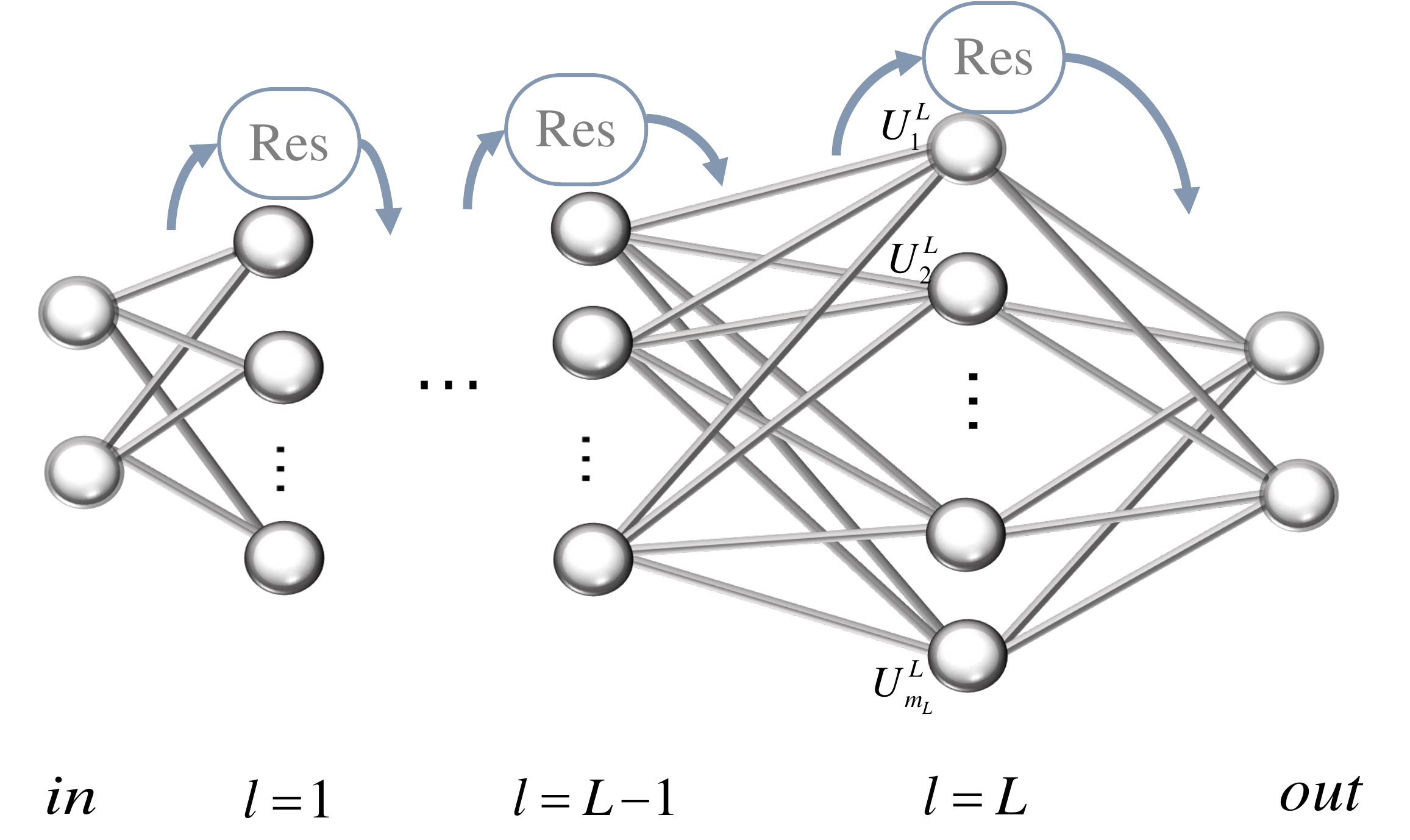}
\caption{\leftskip=0pt \rightskip=0pt plus 0cm  \textbf{The architecture of Res-HQCNN with $L$ hidden layers}. ``Res'' represents the residual block structure of Res-HQCNN in Fig. \ref{fig3}. Not only can the ``Res'' be connected layer by layer continuously, but it can be connected by skipping one or more layers. The architecture of Res-HQCNN propagates information from input to output and gradually goes through a quantum feedforward neural network.}
\label{fig4}
\end{figure}

 Next, we go on investigating the method to merge the residual block structure and quantum neural network together. Define quantum perceptron in layer $l$ of Res-HQCNN to be an arbitrary unitary operator with $m_{l-1}$ input qubits and one output qubit. For example, as presented in Fig. \ref{fig4}, quantum perceptron $U_j^L$ is a $(m_{L-1}+1)$-qubit unitary operator for $j=1,2,\cdots,m_L$. The Res-HQCNN is made up of quantum perceptrons with $L$ hidden layers. It acts on an input state $\rho^{1_{in}}$ of input qubits and obtains a mixed state $\rho^{{L+1}_{out}}$ for the output qubits based on the layer unitary operator $U^{l}$ in the form of a matrix product of quantum perceptrons:
 $U^{l}=U_{m_l}^{l}U_{m_l-1}^{l}\cdots U_{1}^{l}.$ 
 Here $U_{j}^{l}$ acts on the qubits in layer $l-1$ and $l$ for $j=1,2,\cdots,m_l$. The unitary operators are arbitrary, and they do not always commute, so the order of the layer unitary is important. During processing information from $\rho^{1_{in}}$ to $\rho^{{L+1}_{out}}$, the residual block structure produces the new input state for layer $l+1$ through adding the input state with the output state of layer $l$ for $l=1,2,\cdots,L$.

\begin{figure}[ht]
\centering
\includegraphics[angle=0,width=0.7\linewidth]{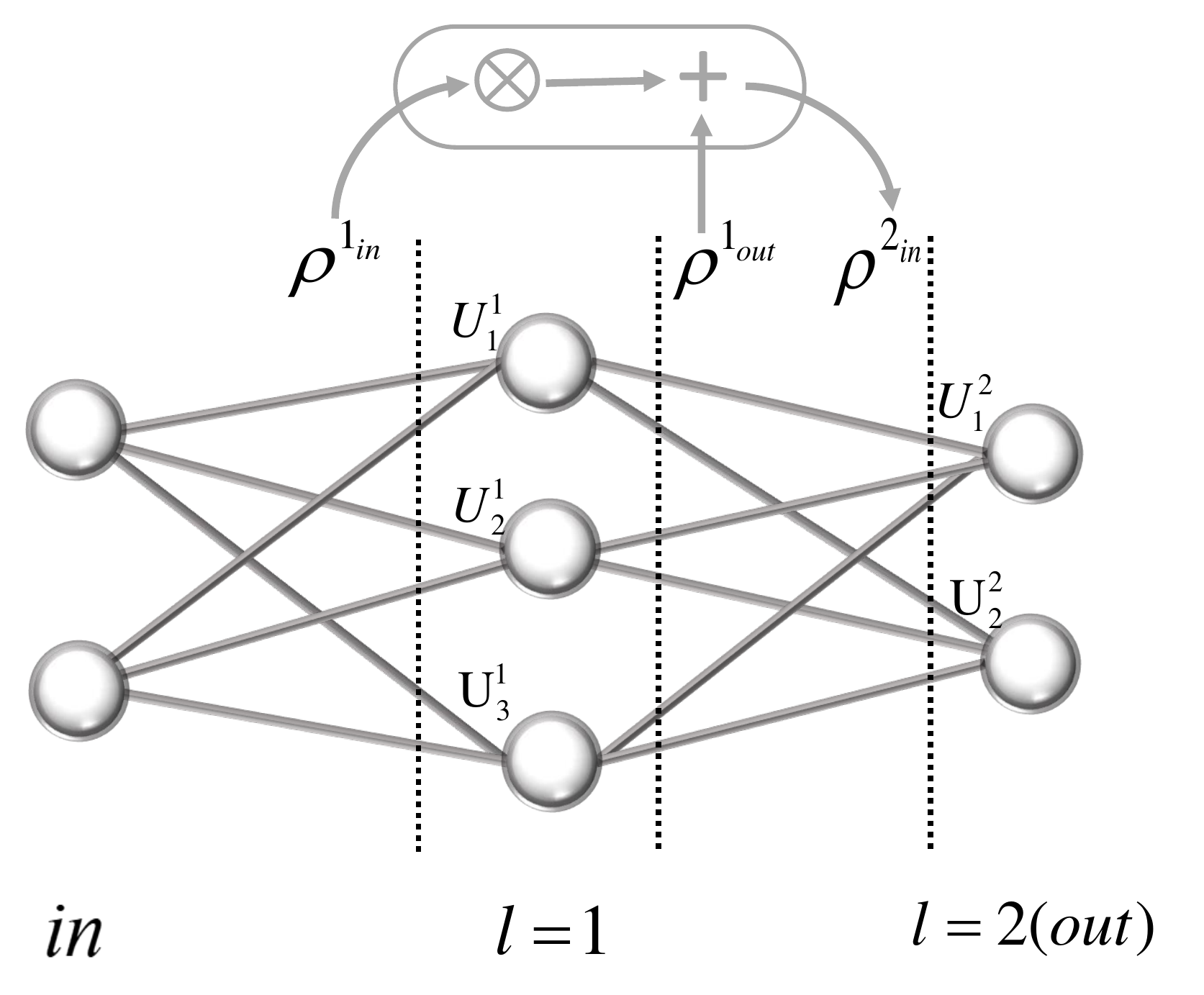}
\caption{ \textbf{The architecture of Res-HQCNN with one hidden layer.} ``$\otimes$'' represents the tensor product of $\rho^{1_{in}}$ and $\ket{0}\bra{0}$. ``$+$'' corresponds to the matrix addition of $\rho^{1_{in}}\otimes \ket{0}\bra{0}$ and $\rho^{1_{out}}$. }
\label{fig5}
\end{figure}

 To facilitate the understanding, we give an example of the mechanism for Res-HQCNN with one hidden layer, see Fig. \ref{fig5}.
 
Define the layer unitary between the input layer and the hidden layer as $U^{1}=U_3^{1}U_2^{1}U_{1}^{1}$, which is in the form of a matrix product of quantum perceptrons. Analogously we define the layer unitary $U^{2}=U_2^{2}U_{1}^{2}$ between the hidden layer and the output layer. For the first step, we apply the quantum perceptrons layer-wise from top to bottom, then the output state $\rho^{1_{out}}$ of the hidden layer is
$$\rho^{1_{out}}=\trace_{in}(U^{1}(\rho^{1_{in}}\otimes\ket{000}_{hid}\bra{000}){U^{1}}^{\dagger}).$$
Next, we apply residual block structure to $\rho^{1_{in}}$ and $\rho^{1_{out}}$ in order to get a new input state for the output layer:
$$\rho^{2_{in}}=\rho^{1_{out}}+\left(\rho^{1_{in}}\otimes \ket{0}\bra{0}\right),$$
In the third step, we get the final output state for this Res-HQCNN in Fig. \ref{fig5}:
$$\rho^{2_{out}}=\trace_{hid}(U^{2}(\rho^{2_{in}}\otimes
\ket{00}_{out}\bra{00}){U^{2}}^{\dagger}).$$

Compared the former QNNs in \cite{beer2020training} with Res-HQCNN, we find that the trace value of the input state 
 $\rho^{{l+1}_{in}}$ for some $l$ changes due to the addition operation in the residual block structure. For example, if we set $L=2$, $\rho^{2_{in}}=\left(\rho^{1_{in}}\otimes \ket{0}_{m_1-m_0}\bra{0}\right)+\rho^{1_{out}}$, and $\rho^{3_{in}}=\left(\rho^{2_{in}}\otimes \ket{0}_{m_2-m_1}\bra{0}\right)+\rho^{2_{out}}$, then the trace value of $\rho^{2_{in}}$ and $\rho^{3_{in}}$ are 2 and 4, respectively. In theory, $\rho^{2_{in}}$ and $\rho^{3_{in}}$ are not density matrix, and we can not apply the training algorithm in quantum computer. However, every coin has two sides. The residual block structure improve the performance of cost function, especially for deeper network, which can be demonstrated in the experiment part.

One may also notice that we can apply the residual block structure for all the hidden layers, but not for the last output layer. Since we have assumed $m_{l-1}\leq m_l$ for $l=1,2,...,L$ and $m_0=m_{L+1}$, then in general, the qubits in the last output layer $L+1$ is no more than the qubits in layer $L$. If we use the residual block structure to the last output layer $L+1$, then the final output of the network will be $\rho^{out}=\rho^{{L+1}_{in}}+\rho^{{L+1}_{out}}$. The dimension of $\rho^{{L+1}_{in}}$ is no less than the dimension of $\rho^{{L+1}_{out}}$, so we should apply partial trace to $\rho^{{L+1}_{in}}$ in order to keep the rule of matrix addition. As we mentioned before, the residual block structure in Res-HQCNN also has difficulty in approximating identity mapping. However, we will lose some information of $\rho^{{L+1}_{in}}$ in this way, which is contradicted with the goal of using residual scheme. We also provide an experiment result to show the inefficiency when applying residual block structure to the last output layer.

\subsection{ The training algorithm of Res-HQCNN}
We randomly generalize $N$ pairs training data which are possibly unknown quantum states in the form of $(\ket{\phi_{x}^{in}}, \ket{\phi_{x}^{out}})$ with $x=1,2,\cdots,N$. It is also allowed to use enough copies of training pair $(\ket{\phi_{x}^{in}}, \ket{\phi_{x}^{out}})$ of specific $x$ so that we can overcome quantum projection noise when computing the derivative of the cost function. Here
for simplicity, we do not allow input states interacting with environment to produce output states (e.g., thermalization). We choose to consider the desired output $\ket{\phi_{x}^{out}}$ as $\ket{\phi_{x}^{out}}= V\ket{\phi_{x}^{in}}$ with $V$ an unknown unitary operation.

The cost function we choose is based on the fidelity between the output of Res-HQCNN and the desired output averaged over all training data. But due to the definition of residual block structure and linear feature of fidelity, we define that the cost function of Res-HQCNN should divide $2^t$, where $t$ is the number of residual block structure in Res-HQCNN:
$$C(s)=\frac{1}{2^t N}\sum_{x=1}^{N}\bra{\phi_{x}^{out}}\rho_{x}^{out}(s)\ket{\phi_{x}^{out}}.$$
Since we want to know how close the network output state and the desired output state, and the closer they are, the bigger fidelity is. If the cost function comes to 1, we judge the Res-HQCNN performs best, otherwise 0 the worst. So
our goal is to maximize the cost function in the training process.

For each layer $l$ of Res-HQCNN, denote $\rho_{x}^{l_{in}}$ as the input state of layer $l$ and $\rho_{x}^{l_{out}}$ as the output state of layer $l$ with $l=1,2,\cdots,L$ and $x=1,2,\cdots,N$. We firstly consider the case that each layer $l$ is added with a residual block structure with no skipping layer, then $t=L$.
The training algorithm for this kind of Res-HQCNN is given by the following steps:
\begin{enumerate}
\item[$\mathbf{I}$.] Initialize:
\begin{enumerate}
    \item[$\mathbf{I1.}$]Set step $s=0$.
    \item[$\mathbf{I2.}$]Choose all unitary $U_{j}^{l}(0)$ randomly, $j=1,2,\cdots,m_l$, where $m_l$ is the number of nodes in layer $l$.
\end{enumerate}

\item[$\mathbf{II}$.] For each layer $l$
and each training pair $(\ket{\phi_{x}^{in}}, \ket{\phi_{x}^{out}})$, do the following steps:
\begin{enumerate}
\item[$\mathbf{II1.}$] Feedforward:
\begin{enumerate}
\item[$\mathbf{II1a.}$] Tensor the input state $\rho_{x}^{l_{in}}(s)$ to the initial state of layer $l$,
$$\rho_{x}^{l_{in}}(s)\otimes \ket{0\cdots0}_{l}\bra{0\cdots0}.$$
Here $\rho_{x}^{1_{in}}(s)=\ket{\phi_{x}^{in}}\bra{\phi_{x}^{in}}$.
\item[$\mathbf{II1b.}$] Apply the layer unitary between layer $l-1$ and $l$,
\begin{align*}
    U_{apply}^{l}(s)=&U_{m_l}^{l}(s)U_{m_l-1}^{l}(s)\cdots U_{1}^{l}(s)(\rho_{x}^{l_{in}}(s)\otimes\nonumber\\ &\ket{0\cdots0}_{l}\bra{0\cdots0})
    {U_{1}^{l}}^{\dagger}(s)\cdots {U_{m_l-1}^{l}}^{\dagger}(s){U_{m_l}^{l}}^{\dagger}(s).
\end{align*}

\item[$\mathbf{II1c.}$]Trace out layer $l-1$ and obtain the output state $\rho_{x}^{l_{out}}(s)$ of layer $l$,
$$\rho_{x}^{l_{out}}(s)=\trace_{l-1}\left(U_{apply}^{l}(s)\right).$$
\end{enumerate}
\item[$\mathbf{II2.}$]Residual learning:
\begin{enumerate}
\item[$\mathbf{II2a.}$] Apply the residual block structure in Fig. \ref{fig3} to $\rho_{x}^{l_{in}}(s)$ and $\rho_{x}^{l_{out}}(s)$ to obtain the new input state of layer $l+1$,
$$\rho_{x}^{l+1_{in}}(s)=\rho_{x}^{l_{out}}(s)+\left(\rho_{x}^{l_{in}}(s)\otimes\ket{0\cdots0}_{\Delta m_l}\bra{0\cdots0}\right).$$
Here $\Delta m_l=m_l-m_{l-1}$ is the number of qubits in $\ket{0\cdots0}$.
\item[$\mathbf{II2b.}$]Store $\rho_{x}^{l+1_{in}}(s)$.

\end{enumerate}
\end{enumerate}

\item[$\mathbf{III}$.] Update parameters:
\begin{enumerate}
\item[$\mathbf{III1.}$]Compute the cost function:
$$C(s)=\frac{1}{2^LN}\sum_{x=1}^{N}\bra{\phi_{x}^{out}}\rho_{x}^{out}(s)\ket{\phi_{x}^{out}}.$$
\item[$\mathbf{III2.}$]Update the unitary of each perceptron via
\begin{align}\label{update}
 U_{j}^{l}(s+\epsilon)=e^{i\epsilon K_{j}^{l}(s)}U_{j}^{l}(s).
\end{align}
Here $K_{j}^{l}(s)$ is the parameters matrix.
\begin{enumerate}
\item[$\mathbf{III2a.}$]
If Res-HQCNN has three layers with one hidden layer.

When $l=1$, we can get the analytical expression for $K_{j}^{l}$ after calculation,
\begin{align}\label{k1a}
K_{j}^{l}(s)=\eta \frac{2^{m_{l-1}}}{N}\sum_{x=1}^{N}\trace_{rest}M_{j}^{l},
\end{align}
where the trace is over all qubits of Res-HQCNN which are not affected by $U_{j}^{l}$. $\eta$ is the learning rate and N is the number of training pairs.
Moreover, $M_{j}^{l}$ is made up of two parts of the commutator:
\begin{align}\label{k1}
 M_{j}^{l}(s)=&[U_{j}^{l}(s)\cdots U_{1}^{l}(s)\left(\rho_{x}^{l_{in}}(s)\otimes \ket{0\cdots0}_{l}\bra{0\cdots0}\right){U_{1}^{l}}^{\dagger}(s)\cdots 
 \nonumber\\
&{U_{j}^{l}}^{\dagger}(s),{U_{j+1}^{l}}^{\dagger}(s)\cdots {U_{m_{out}}^{out}}^{\dagger}(s)
\left(Id(m_{l-1})\otimes \ket{\phi_{x}^{out}}\bra{\phi_{x}^{out}}\right)\nonumber\\
&U_{m_{out}}^{out}(s)\cdots U_{j+1}^{l}(s)].
\end{align}

When $l=2$, then 
\begin{align}\label{k2a}
K_{j}^{l}(s)=\eta \frac{2^{m_{l-1}}}{N}\sum_{x=1}^{N}\trace_{rest}(M_{j}^{l}+N_{j}^{l}),
\end{align}
where $M_{j}^{l}$ is Eq.(\ref{k1}) and
\begin{align}\label{K2}
 N_{j}^{2}(s)=&[U_{j}^{2}(s)\cdots U_{1}^{2}(s)(\rho_{x}^{in}(s)\otimes \ket{0\cdots0}_{\Delta m_1}\bra{0\cdots0}\otimes\nonumber\\ &\ket{0\cdots0}_{2}\bra{0\cdots0})
 {U_{1}^{2}}^{\dagger}(s)\cdots {U_{j}^{2}}^{\dagger}(s),
{U_{j+1}^{2}}^{\dagger}(s)\cdots {U_{m_{2}}^{2}}^{\dagger}(s)
\nonumber\\
&\left(Id(m_{1})\otimes \ket{\phi_{x}^{out}}\bra{\phi_{x}^{out}}\right)U_{m_{2}}^{2}(s)\cdots U_{j+1}^{2}(s)].
\end{align}
\item[$\mathbf{III2b.}$]
If Res-HQCNN has four layers with two hidden layers.

When $l=1$, then 
\begin{align}\label{k3a}
K_{j}^{l}(s)=\eta \frac{2^{m_{l-1}}}{N}\sum_{x=1}^{N}\trace_{rest}(M_{j}^{l}+P_{j}^{l}),
\end{align}
where $M_{j}^{l}$ is Eq.(\ref{k1}) and
\begin{align}\label{k3}
 P_{j}^{1}(s)=&[U_{j}^{1}(s)\cdots U_{1}^{1}(s)\left(\rho_{x}^{in}(s)\otimes \ket{0\cdots0}_{1}\bra{0\cdots0}\right)
{U_{1}^{1}}^{\dagger}(s)\cdots  \nonumber\\
& {U_{j}^{1}}^{\dagger}(s), {U_{j+1}^{1}}^{\dagger}(s)\cdots{U_{m_{1}}^{1}}^{\dagger}(s){U_1^{3}}^{\dagger}(s)\cdots {U_{m_{3}}^{3}}^{\dagger}(s)(Id(m_{2})\nonumber\\
&\otimes \ket{\phi_{x}^{out}}\bra{\phi_{x}^{out}})U_{m_{3}}^{3}\cdots U_1^{3}U_{m_{1}}^{1}(s)\cdots U_{j+1}^{1}(s)].
\end{align}

When $l=2$, then 
\begin{align}\label{k4a}
K_{j}^{l}(s)=\eta \frac{2^{m_{l-1}}}{N}\sum_{x=1}^{N}\trace_{rest}(M_{j}^{l}+Q_{j}^{l}),
\end{align}
where $M_{j}^{l}$ is Eq.(\ref{k1}) and
\begin{align}\label{k4}
 Q_{j}^{2}(s)=&[U_{j}^{2}(s)\cdots U_{1}^{2}(s)(\rho_{x}^{in}(s)\otimes \ket{0\cdots0}_{\Delta m_1}\bra{0\cdots0}\otimes \nonumber\\
 &\ket{0\cdots0}_{2}\bra{0\cdots0}){U_{1}^{2}}^{\dagger}(s)\cdots{U_{j}^{2}}^{\dagger}(s),{U_{j+1}^{2}}^{\dagger}(s)\cdots {U_{m_{2}}^{2}}^{\dagger}(s)
\nonumber\\
& {U_1^{3}}^{\dagger}(s)\cdots {U_{m_{3}}^{3}}^{\dagger}(s)\left(Id(m_{2})\otimes \ket{\phi_{x}^{out}}\bra{\phi_{x}^{out}}\right)U_{m_{3}}^{3}\cdots U_1^{3}\nonumber\\
&U_{m_{2}}^{2}(s)\cdots U_{j+1}^{2}(s)].
\end{align}

When $l=3$, then 
\begin{align}\label{k5a}
K_{j}^{l}(s)=\eta \frac{2^{m_{l-1}}}{N}\sum_{x=1}^{N}\trace_{rest}(M_{j}^{l}+S_{j}^{l}+T_{j}^{l}),
\end{align}
where $M_{j}^{l}$ is Eq.(\ref{k1}) and
\begin{align}\label{k5}
 S_{j}^{3}(s)=&[U_{j}^{3}(s)\cdots U_{1}^{3}(s)(\rho_{x}^{in}(s)\otimes \ket{0\cdots0}_{\Delta m_1+\Delta m_2}\bra{0\cdots0}\nonumber\\
 &\otimes \ket{0\cdots0}_{3}\bra{0\cdots0}){U_{1}^{3}}^{\dagger}(s)\cdots {U_{j}^{3}}^{\dagger}(s),
{U_{j+1}^{3}}^{\dagger}(s)\cdots 
\nonumber\\
&{U_{m_{3}}^{3}}^{\dagger}(s)\left(Id(m_{2})\otimes \ket{\phi_{x}^{out}}\bra{\phi_{x}^{out}}\right)U_{m_{3}}^{3}\cdots U_{j+1}^{3}(s)],
\end{align}

\begin{align}\label{k6}
 T_{j}^{3}(s)=&[U_{j}^{3}(s)\cdots U_{1}^{3}(s)U_{m_1}^{1}(s)\cdots U_{1}^{1}(s)(\rho_{x}^{in}(s)\otimes  \nonumber\\
&\ket{0\cdots0}_{1}\bra{0\cdots0}){U_{1}^{1}}^{\dagger}(s)\cdots {U_{m_1}^{1}}^{\dagger}(s){U_{1}^{3}}^{\dagger}(s)\cdots {U_{j}^{3}}^{\dagger}(s),
\nonumber\\
&{U_{j+1}^{3}}^{\dagger}(s)\cdots {U_{m_{3}}^{3}}^{\dagger}(s)\left(Id(m_{2})\otimes \ket{\phi_{x}^{out}}\bra{\phi_{x}^{out}}\right)U_{m_{3}}^{3}\cdots U_{j+1}^{3}(s)].
\end{align}
\item[$\mathbf{III2c.}$]
If Res-HQCNN has more than five layers,
we can calculate the corresponding parameters matrices $K_{j}^{l}(s)$ with the method in Appendix. There is no fixed formula for $K_{j}^{l}(s)$. It changes with the depth of network and the parameter $l$.
\end{enumerate}
\item[$\mathbf{III3.}$] Update $s=s+\epsilon$.
\end{enumerate}

\item[$\mathbf{IV}$.] Repeat steps $\mathbf{II}$ and $\mathbf{III}$ until reaching the maximum of the cost function.
\end{enumerate}

As for other cases of Res-HQCNN, we can get $t\leq L$ due to the skipping connection. The calculation method is the same as the one in Appendix, and the complexity of calculation decrease compared with the case $t=L$. So we do not give the training algorithm for other cases in detail. 
Here we can automatically update the unitary of each perceptron using Eq. (\ref{update}) until converging to the optimal value of the cost function. Therefore, we can conclude that this Res-HQCNN can be trained in an end-to-end fashion.

Compared with the training algorithm on classical computer of QNNs in \cite{beer2020training}, we find that $K_{j}^{l}(s)$ in QNNs is no more than the one in Res-HQCNN. The residual block structure brings the new items for $K_{j}^{l}(s)$, such as $N_{j}^{l}(s)$, $P_{j}^{l}(s)$, $Q_{j}^{l}(s)$, $S_{j}^{l}(s)$ and $T_{j}^{l}(s)$ in Eq. (\ref{K2}, \ref{k3}, \ref{k4}, \ref{k5}, \ref{k6}). Observing the analytical expressions of Eq. (\ref{K2}, \ref{k3}, \ref{k4}, \ref{k5}, \ref{k6}) in detail, we notice that the input state can be applied by any deeper layer from the perspective of propagating information feedforward (e.g. $N_{j}^{l}(s)$, $Q_{j}^{l}(s)$, $S_{j}^{l}(s)$, $T_{j}^{l}(s)$) and backward (e.g. $P_{j}^{l}(s)$).

For better understanding of the training algorithm, we also provide a simple flowchart for the readers as following:
\begin{figure}[ht]
\centering
\includegraphics[angle=0,width=0.7\linewidth]{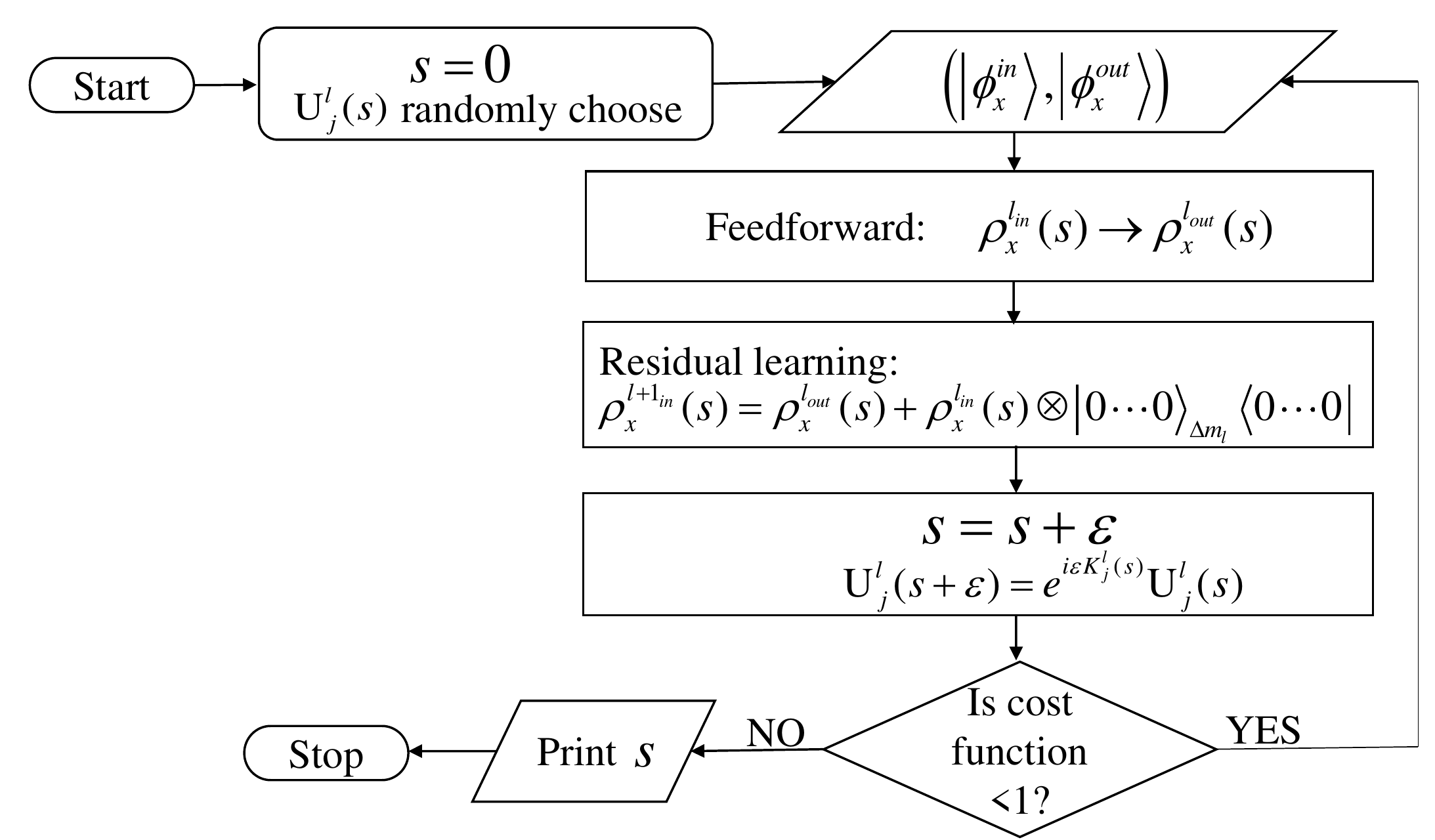}
\caption{ \textbf{Flowchart of training algorithm when $t=L$.} }
\label{fig6}
\end{figure}

\section{Experiments}\label{5}
In this section, we conduct comprehensive experiments to evaluate the performance of Res-HQCNN. The experiments were run  on ThinLinc server with 2 x Intel Xeon CPU E5-2650 v3 (2.30GHz)  on CentOS (Linux RHEL clone) Operating system.
Firstly, we start from the elementary tests to prove the effectiveness. Here we also show the experimental results if we apply residual block structure to the last output layer. Then, we further explore the performance of Res-HQCNN with more layers. For four-layer Res-HQCNN, we compare the performance of different types, such as skipping one layer to connect the residual block structure. Finally, we generalize the experiments into noisy training data for testing the robustness of Res-HQCNN. 

The training data for the following elementary tests and big networks are possibly unknown quantum states in the form of $(\ket{\phi_x^{in}},V\ket{\phi_x^{in}})$ for $x=1,2,\cdots,N$ with an unknown unitary matrix $V$. The elements of $\ket{\phi_x^{in}}$ are randomly picked out of a normal distribution before normalization. The elements of unitary matrix $V$ are randomly picked out of a normal distribution before orthogonalization. The training data in the form of $(\ket{\phi_x^{in}},V\ket{\phi_x^{in}})$ are regarded as good training data. Compared with good training data, the noisy training data means that the training pairs are in the form of $(\ket{\phi_x^{in}},\ket{\theta_x^{out}})$, where the desired output $\ket{\theta_x^{out}}$ has no direct transform relation to $\ket{\phi_x^{in}}$.
Both good and noisy training data in the experiments are randomly generated quantum states.

In order to show the power of residual block structure in Res-HQCNN, we compare with the results using the training algorithm of QNNs in~\cite{beer2020training}.
For convenience, we still apply a 1-dimensional list of natural numbers to refer to the number of perceptrons in the corresponding layer as in Fig. \ref{Figure1}. Specially, if there is a residual block structure shown in Fig. \ref{fig3} that acts on the hidden layers, we plus a tilde on the top of the natural numbers. For example, a 1-2-1 quantum neural network in \cite{beer2020training} can be denoted as $[1,2,1]$, and a 1-2-1 quantum neural network with our residual block structure in this paper will be written as $[1,\tilde{2},1]$. If there is a skipping connection, we plus a hat on the top of the number representing the skipped layer. For example, $[2,\hat{3},\tilde{3},2]$ denotes $\rho_{x}^{3_{in}}(s)=\rho_{x}^{2_{out}}(s)+\left(\rho_{x}^{1_{in}}(s)\otimes\ket{0}\bra{0}\right).$

\subsection{ Elementary tests}
\begin{figure*}[htbp]
\centering
\subfigure[$\eta=1/1.8$ and $\epsilon=0.1$]{
\includegraphics[width=6cm]{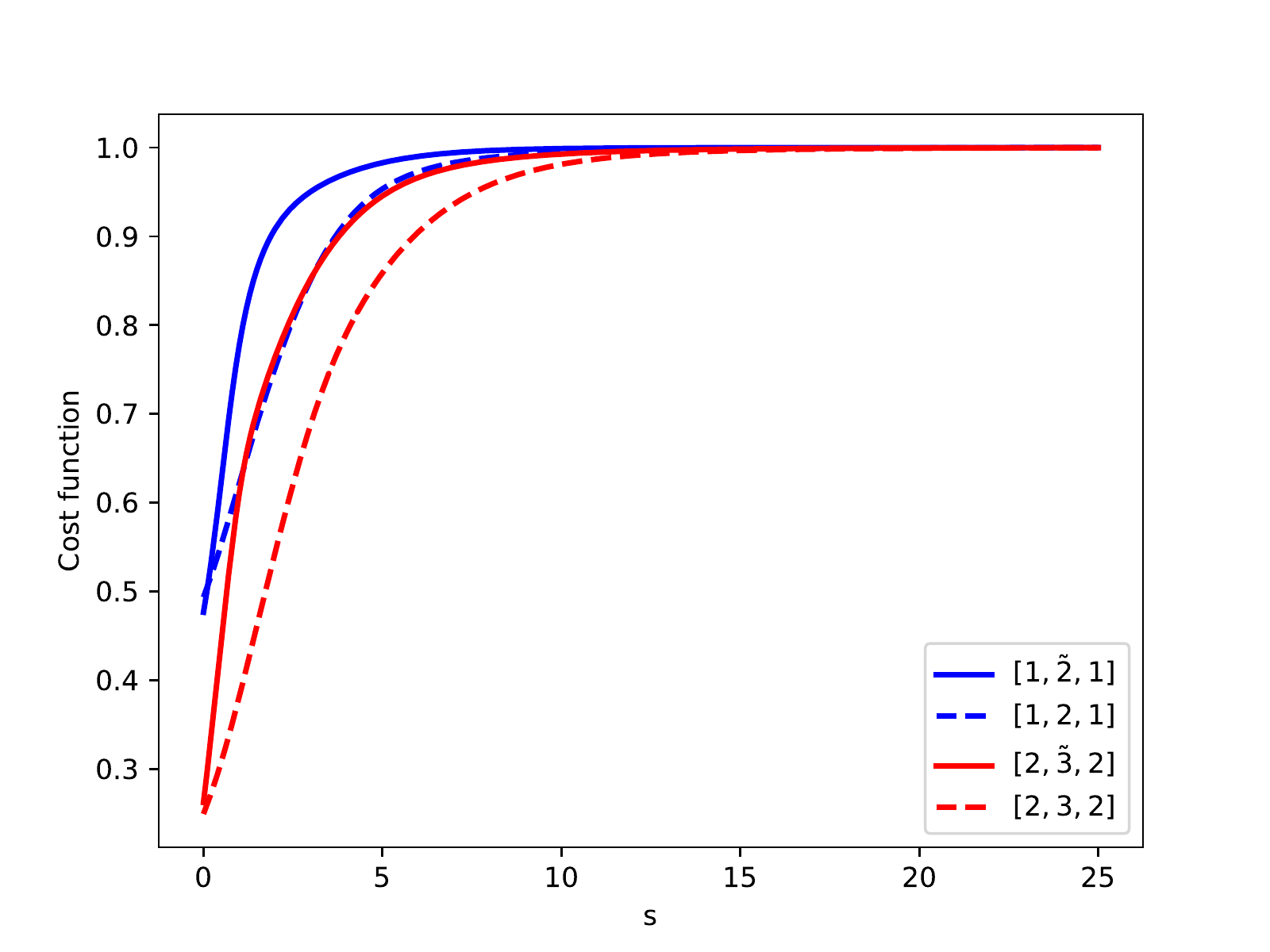}
}
\subfigure[$\eta=1/2$ and $\epsilon=0.1$]{
\includegraphics[width=6cm]{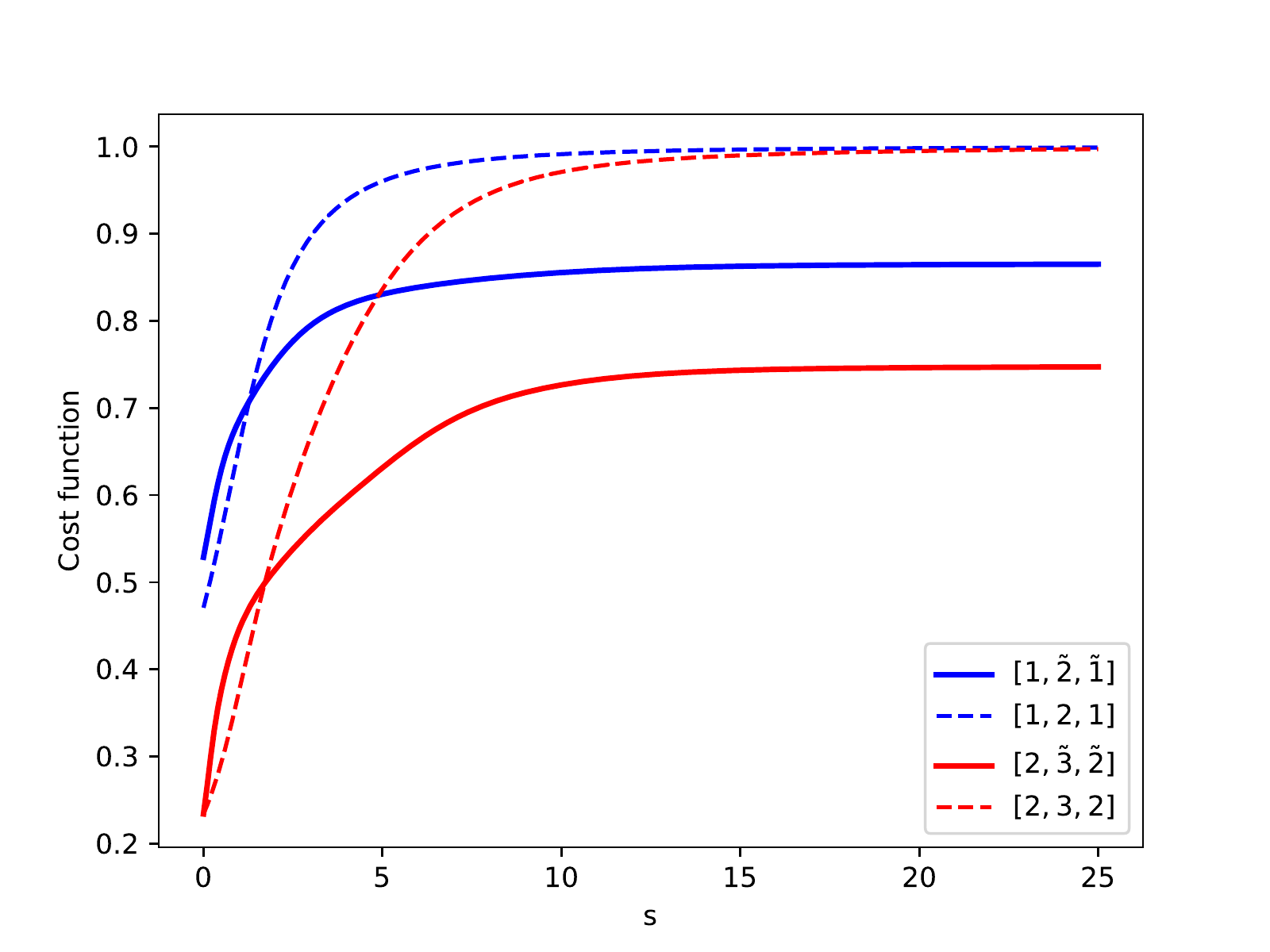}
}
\caption{ \textbf{Numerical results of $[1,\tilde{2},1]$, $[2,\tilde{3},2]$,$[1,\tilde{2},\tilde{1}]$ and $[2,\tilde{3},\tilde{2}]$ with 10 training pairs for 250 training rounds.}}
\label{Figure7}
\end{figure*}
We consider Res-HQCNN $[1,\tilde{2},1]$ and $[2,\tilde{3},2]$ with $\eta=1/1.8$ and $\epsilon=0.1$ for elementary tests, see  Fig. \ref{Figure7}(a). We find that the solid line is higher than the dashed one in the same color and both lines converge to 1 as the training rounds increasing to 250. Compared the solid lines with dashed lines in the same color, we find that the solid lines have higher rate of convergence than the dashed ones before reaching $1$. 

We then test the performance of $[1,\tilde{2},\tilde{1}]$ and $[2,\tilde{3},\tilde{2}]$ with $\eta=1/2$ and $\epsilon=0.1$, which is the case that applying the residual block structure to the last output layer, see Fig. \ref{Figure7}(b). Since the residual block structure are used twice in $[1,\tilde{2},\tilde{1}]$ and $[2,\tilde{3},\tilde{2}]$, the update parameters matrix $K_j^l$ in Eq. (\ref{k2a}) will increase by adding one more term after computation. The method of computation is in the Appendix. So we set $\eta=1/2$ to decrease the learning rate a bit. In Fig. \ref{Figure7}(b), the solid lines in blue and red can not reach the maximum of the cost function. Compared the solid and dashed lines in the same color, the rate of convergence of the solid lines is not always higher than the one of dashed lines. These results give agreement with the theoretical analysis in the last paragraph of Subsection 4.1. Therefore we will not use residual block structure to the last output layer for Res-HQCNN in the following.

\subsection{ Big networks}

\begin{figure*}[htbp]
\centering
\subfigure[$\eta=1/5$ and $\epsilon=0.1$.]{
\includegraphics[width=6cm]{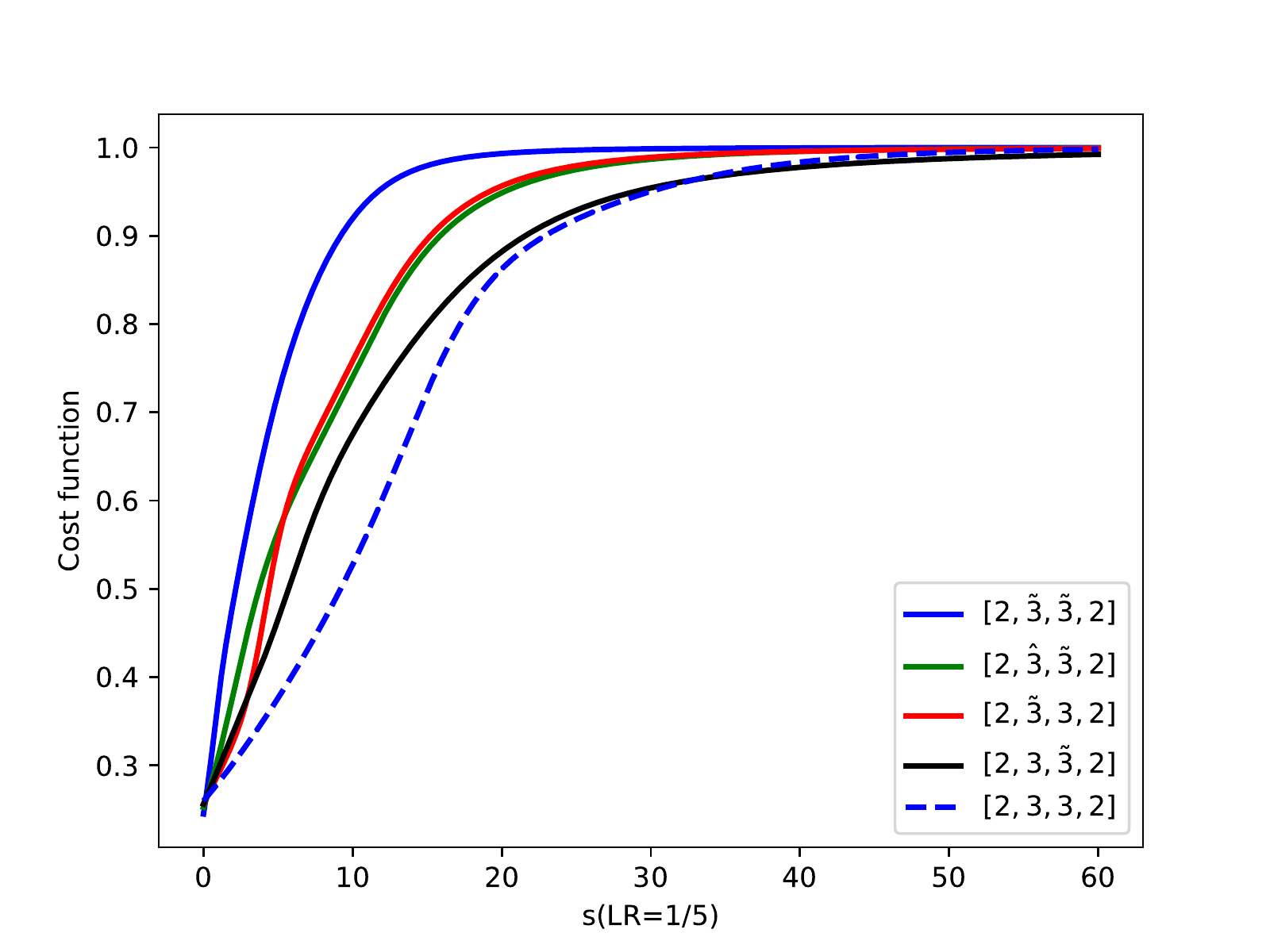}
}
\subfigure[$\eta=1/3$ and $\epsilon=0.1$.]{
\includegraphics[width=6cm]{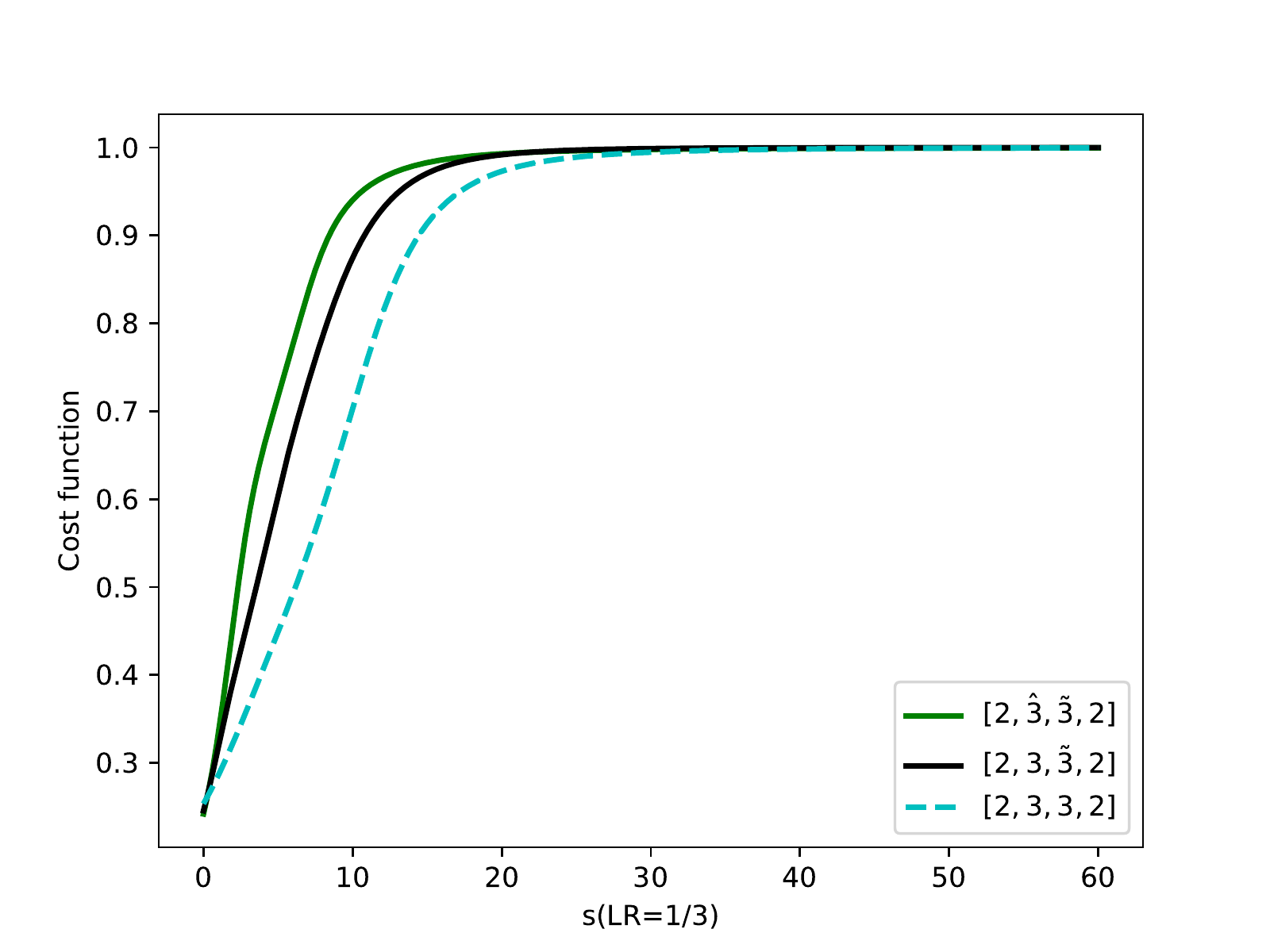}
}
\caption{ \textbf{Numerical results of $[2,3,3,2]$ with residual block structure in different cases for 10 training pairs and 600 training rounds.} }
\label{Figure8}
\end{figure*}

\begin{figure*}[htbp]
\centering
\subfigure[$\eta=1/9$ and $\epsilon=0.1$]{
\includegraphics[width=6cm]{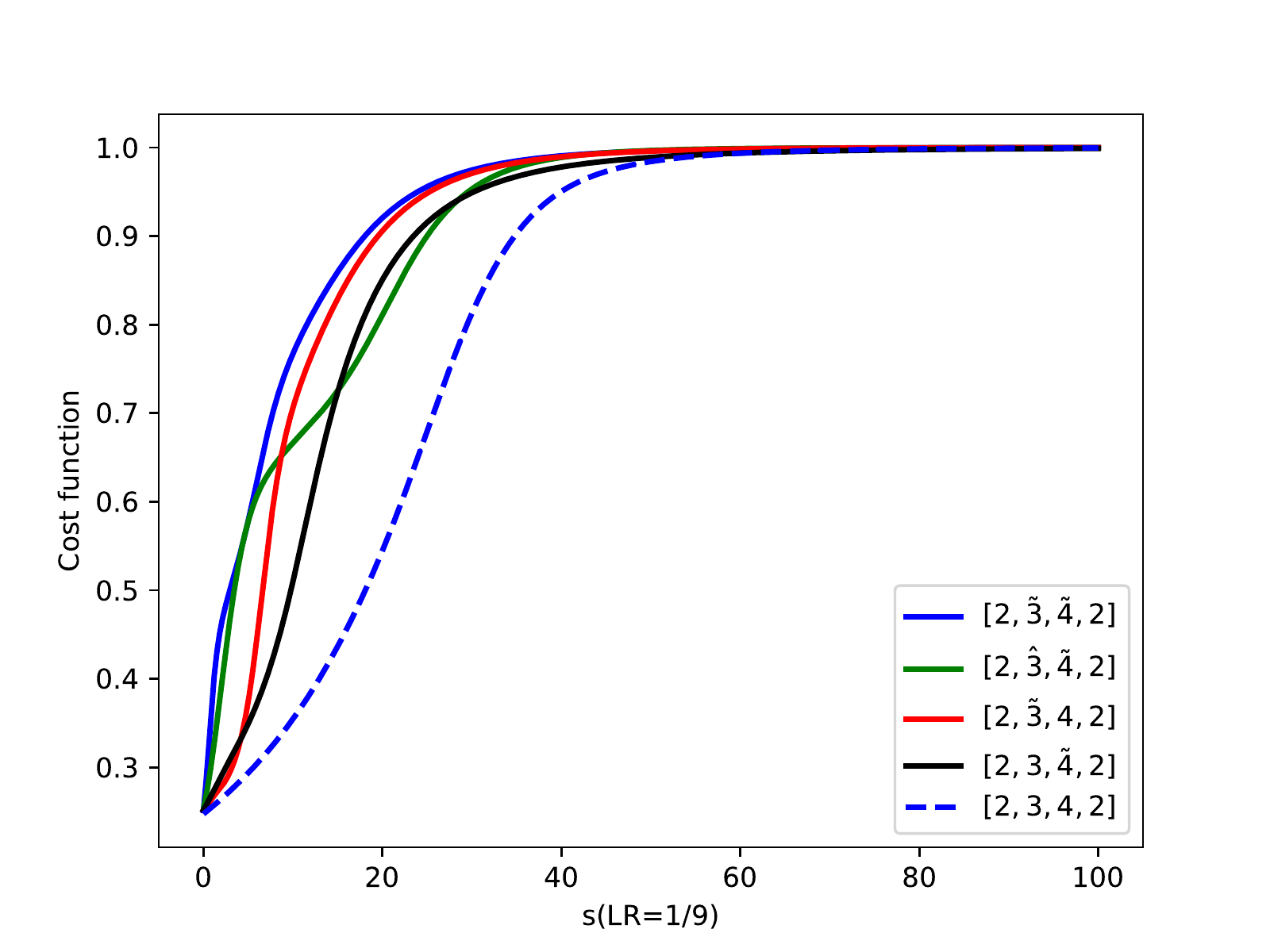}
}
\subfigure[$\eta=1/5$ and $\epsilon=0.1$]{
\includegraphics[width=6cm]{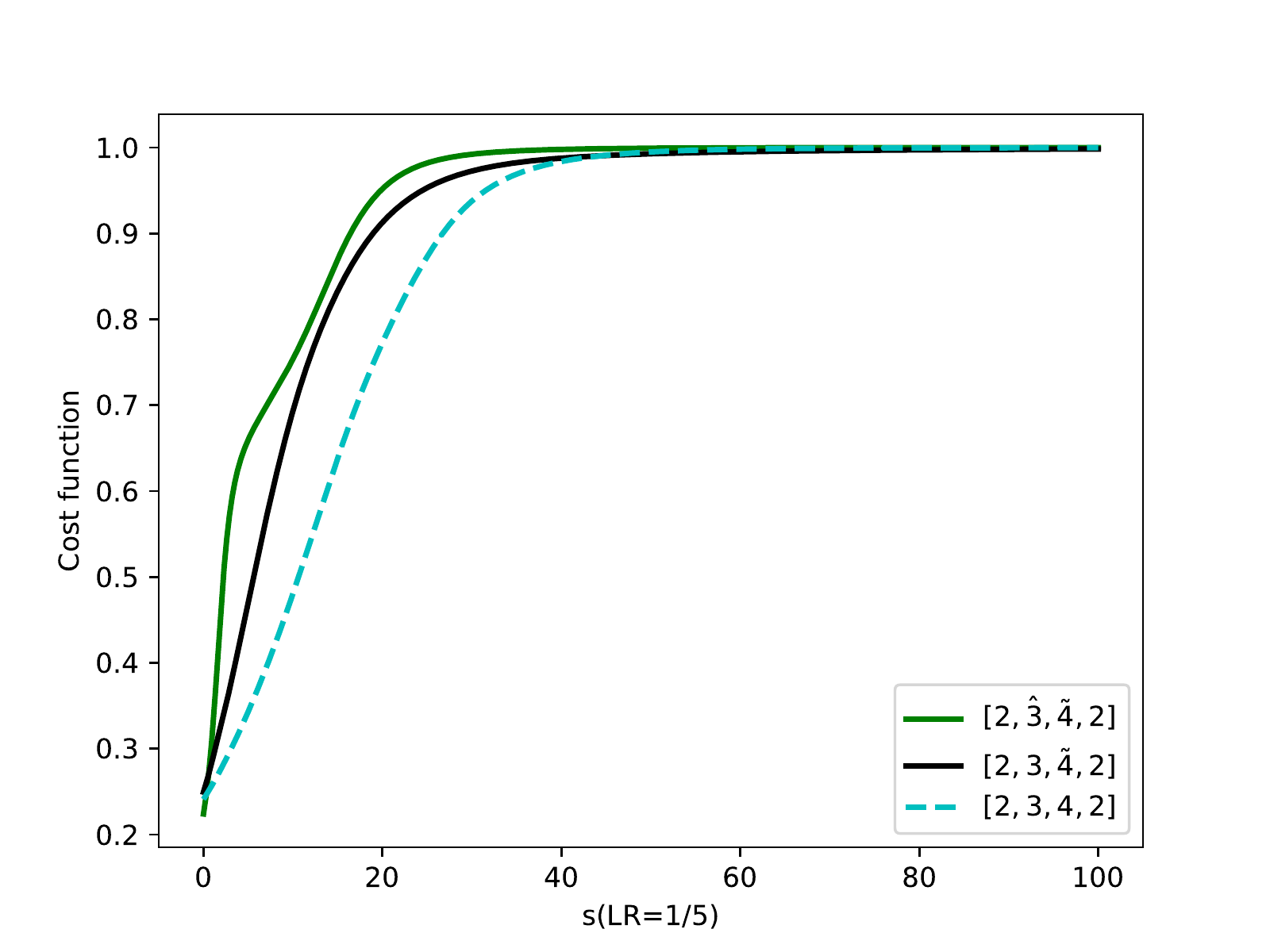}
}
\caption{\textbf{Numerical results of $[2,3,4,2]$ with residual block structure in different cases for 10 training pairs and 1000 training rounds.} }
\label{Figure9}
\end{figure*}

In this subsection, we consider Res-HQCNN with deeper layers to test the advantages of deep residual learning. We firstly select Res-HQCNN $[2,\tilde{3},\tilde{3},2]$ and $[2,\tilde{3},\tilde{4},2]$ for 10 training pairs.
The corresponding simulation results of them are presented in Fig. \ref{Figure8} and Fig. \ref{Figure9}.

In Fig. \ref{Figure8}(a), we test all possible four cases for QNNs $[2,3,3,2]$ with residual block structure. Since $[2,\tilde{3},\tilde{3},2]$ has two residual block structures, the parameters matrix $K_j^l$ for $[2,\tilde{3},\tilde{3},2]$ would be larger than the ones for $[2,\hat{3},\tilde{3},2]$, $[2,3,\tilde{3},2]$ and $[2,\tilde{3},3,2]$. Choosing $\eta=1/5$ is suitable for $[2,\tilde{3},\tilde{3},2]$, but smaller for other cases. So the lines of $[2,\hat{3},\tilde{3},2]$, $[2,3,\tilde{3},2]$ and $[2,\tilde{3},3,2]$ are a bit unstable in Fig. \ref{Figure8}(a). However, it does not matter. From Fig. \ref{Figure8}(a), we can see that all possible four cases for QNNs $[2,3,3,2]$ with residual block structure perform well and $[2,\tilde{3},\tilde{3},2]$ performs best than all other cases.

In order to test the power of skipping connection, we plot the performance of $[2,\hat{3},\tilde{3},2]$ and $[2,3,\tilde{3},2]$ with suitable learning rate $\eta=1/3$ in 
Fig. \ref{Figure8}(b). One can find that $[2,\hat{3},\tilde{3},2]$ with $\rho_{x}^{3_{in}}(s)=\rho_{x}^{2_{out}}(s)+\left(\rho_{x}^{1_{in}}(s)\otimes\ket{0}\bra{0}\right)$ performs better than $[2,3,\tilde{3},2]$ with $\rho_{x}^{3_{in}}(s)=\rho_{x}^{2_{out}}(s)+\rho_{x}^{2_{in}}(s)$. Therefore we find that the skipping connection here is helpful in improving the performance of cost function.

We then try to change QNNs $[2,3,4,2]$ into Res-HQCNN with all possible cases, see Fig. \ref{Figure9}(a). Analogous to the results in Fig. \ref{Figure8}(a), the blue line is above the other lines. The red and green lines are still unstable due to a smaller learning rate. We still test the skipping connection for $[2,\hat{3},\tilde{4},2]$ and $[2,3,\tilde{4},2]$ to see a difference. The green line is higher than the black one, which is consistent with the previous results in Fig. \ref{Figure8}(b).

Taken Fig. \ref{Figure8} and Fig. \ref{Figure9} together, one can read out that if a QNNs with $L$ hidden layers, then the case that Res-HQCNN has $L$ residual block structures can bring the best improvement of cost function than the other cases, including the case with skipping connection. So in the following, we will consider the best case of Res-HQCNN with $L$ residual block structures.

\begin{figure*}[htbp]
\centering
\subfigure[$\eta=1/15$ and $\epsilon=0.1$]{
\includegraphics[width=6cm]{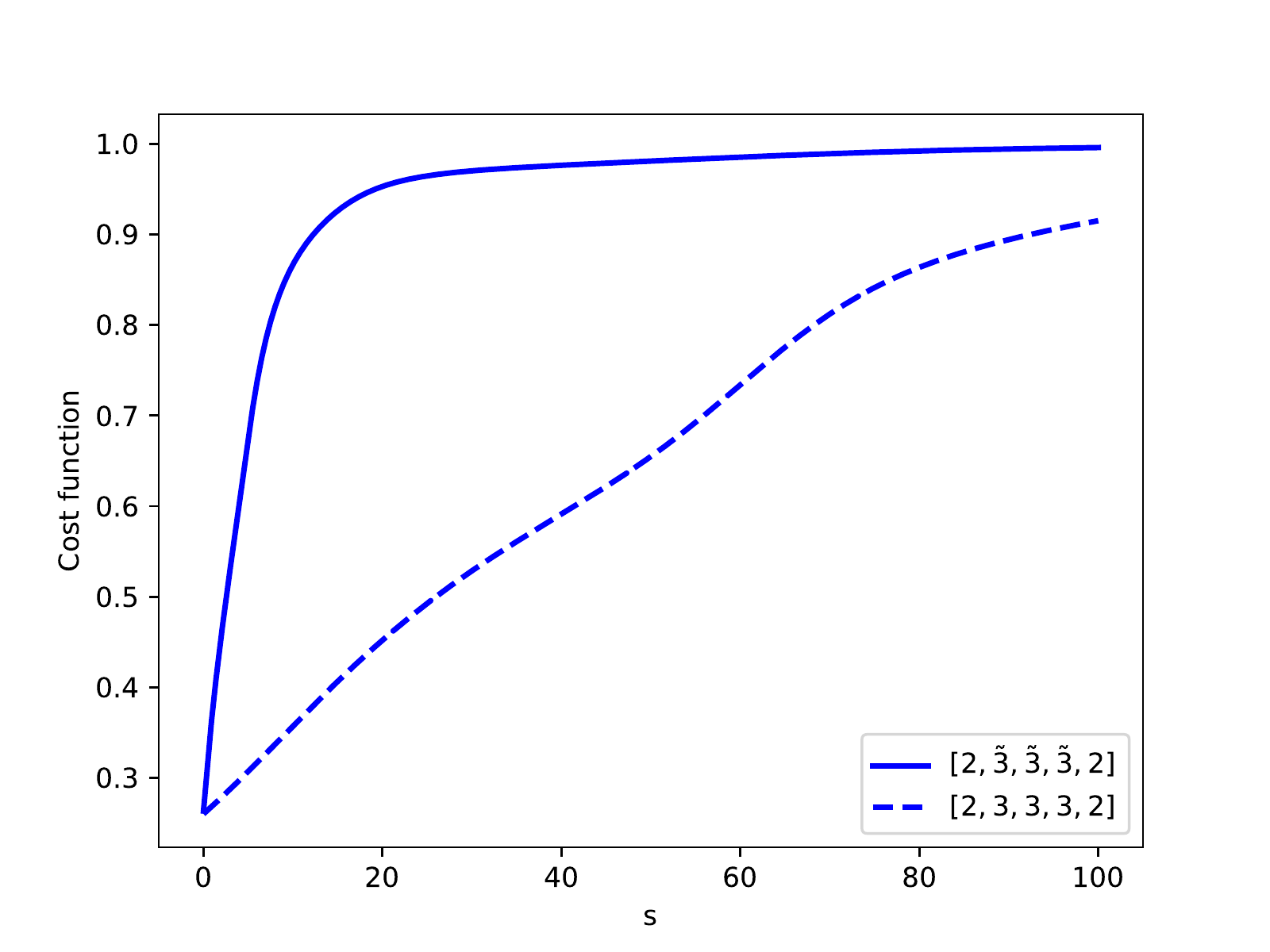}
}
\subfigure[$\eta=1/35$ and $\epsilon=0.1$]{
\includegraphics[width=6cm]{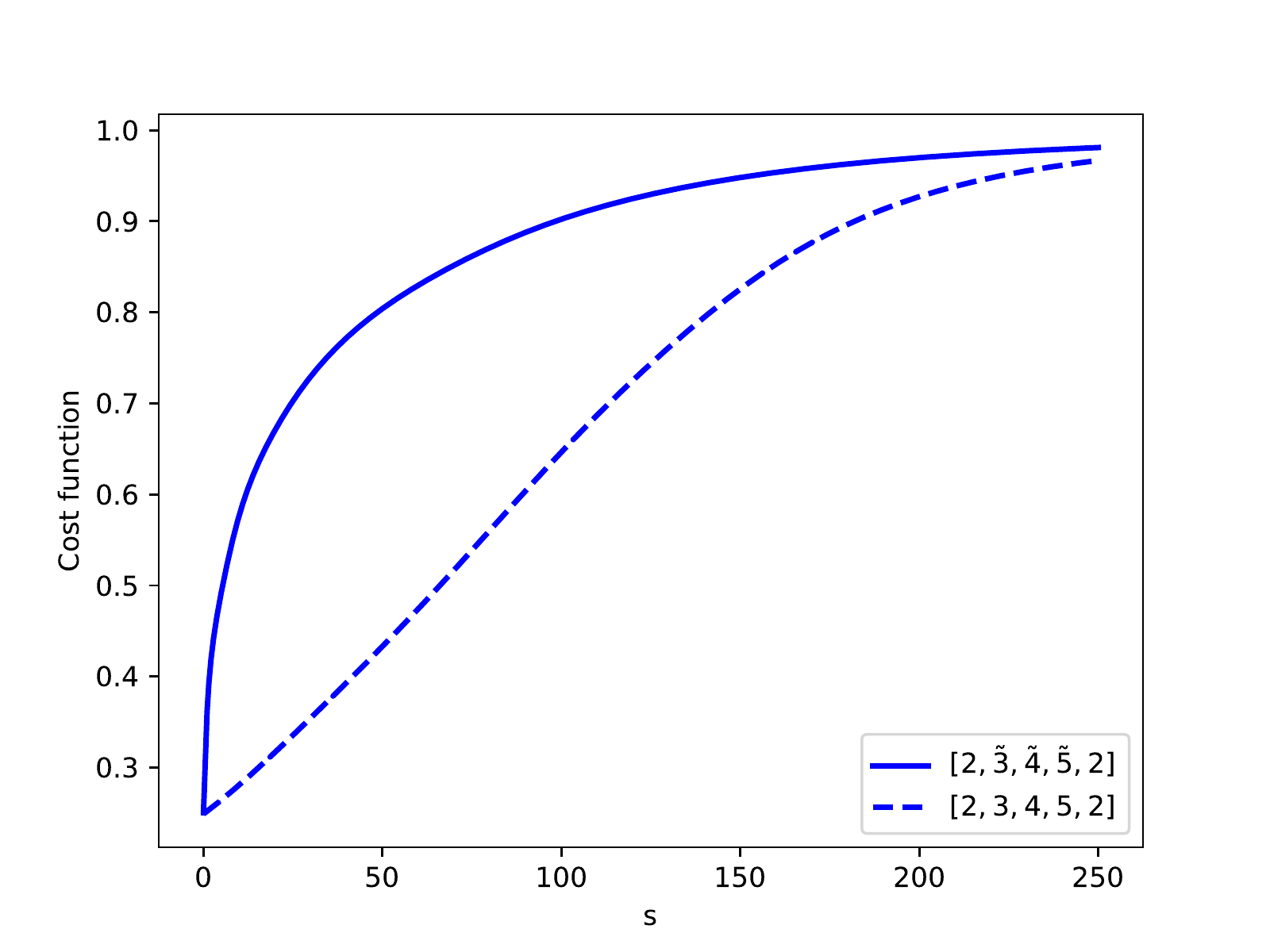}
}
\caption{\textbf{Numerical results of $[2,\tilde{3},\tilde{3},\tilde{3},2]$ and $[2,\tilde{3},\tilde{4},\tilde{5},2]$ with 5 training pairs for 1000 training rounds and 2500 training rounds, respectively.} }
\label{Figure10}
\end{figure*}

In order to test the effectiveness of residual block structure in quantum concept further, we go on detecting deeper Res-HQCNN $[2,\tilde{3},\tilde{3},\tilde{3},2]$ and $[2,\tilde{3},\tilde{4},\tilde{5},2]$. 
When the number of training pairs is set to be 5 with training rounds 1000 in Fig. \ref{Figure10}(a), we find that the solid line is increasing with decreasing slope and finally it converges to $1$ as the training rounds increasing to 600.  The slope of the bottom dashed line is increasing, but the dashed line does not converge at all during 1000 training rounds. This result is quite impressive.
Next in Fig. \ref{Figure10}(b), the cost function of $[2,\tilde{3},\tilde{4},\tilde{5},2]$ converges slower than the one of $[2,\tilde{3},\tilde{3},\tilde{3},2]$ due to the increase of quantum perceptrons. But the value of cost function for $[2,\tilde{3},\tilde{4},\tilde{5},2]$ is always larger than the one for $[2,3,4,5,2]$ and solid line has larger convergence speed. 

Compared Fig. \ref{Figure10} with the former Fig. \ref{Figure7}, Fig. \ref{Figure8} and Fig. \ref{Figure9} together, one can see that deeper Res-HQCNN could bring more significant improvement, such as the difference of convergence rate between the solid and dashed line in the same color. As we know, if a Res-HQCNN has $L$ hidden layers, then it will use $L$ residual block structures to perform well. The larger $L$ is, the bigger the parameters matrix $K_j^l$ is, which results in a smaller suitable learning rate for Res-HQCNN. So when compared the performance of QNNs and Res-HQCNN at small learning rate, we will find out deeper Res-HQCNN learns faster and better than the former QNNs.

\subsection{Generalization: the robustness to noisy data}

In this subsection, we begin examining the robustness of Res-HQCNN to noisy quantum data. For comparison conveniently, we employ the same rule in \cite{beer2020training} to test the robustness.
We firstly generate $N$ good training pairs $(\ket{\phi_x^{in}},V\ket{\phi_x^{in}})$ and then destroy $n$ of them by replacing them with noisy training pairs. Every time the replaced subset is chosen randomly. The cost function is assessed for all good test pairs.

\begin{figure*}[ht]
\centering
\subfigure[30 training pairs and 50 training rounds]{
\includegraphics[angle=0,width=0.7\linewidth]{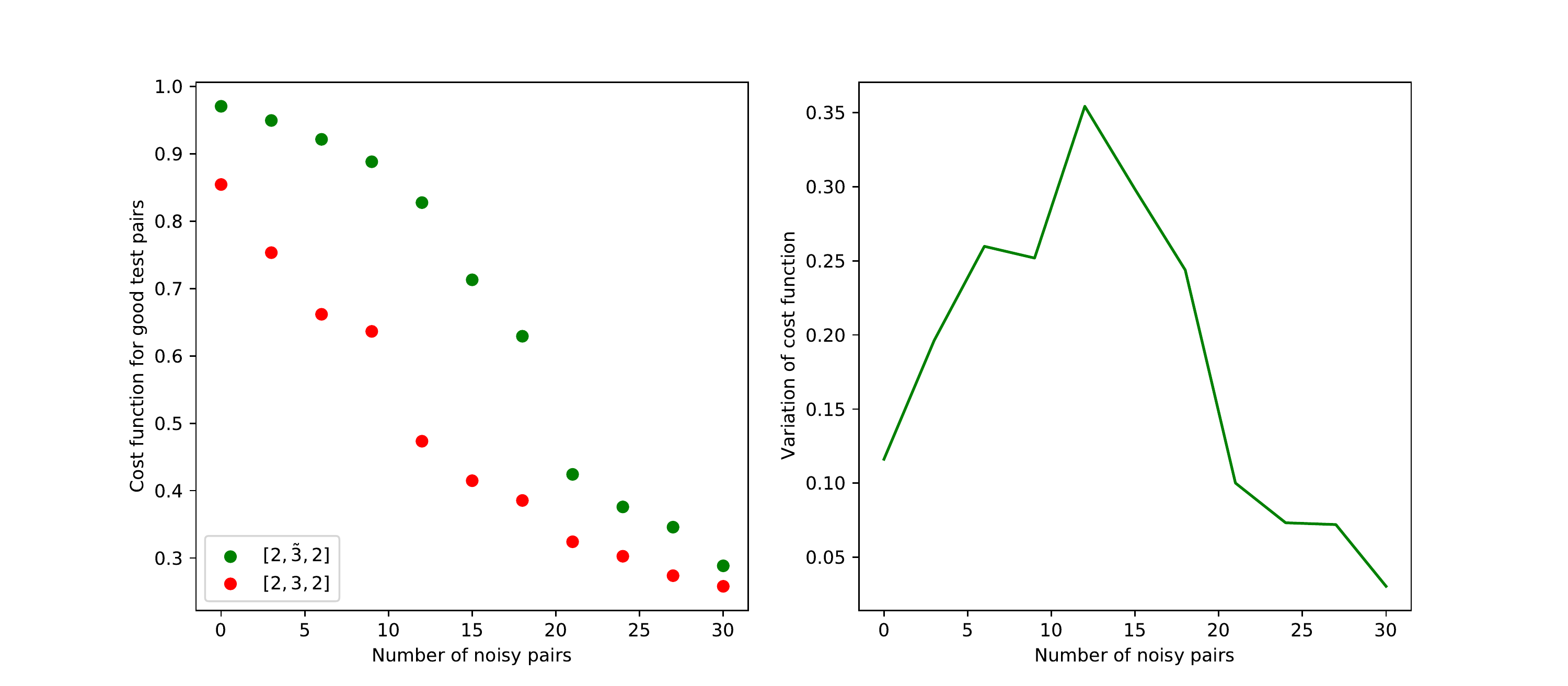}
}
\subfigure[100 training pairs and 200 training rounds]{
\includegraphics[angle=0,width=0.7\linewidth]{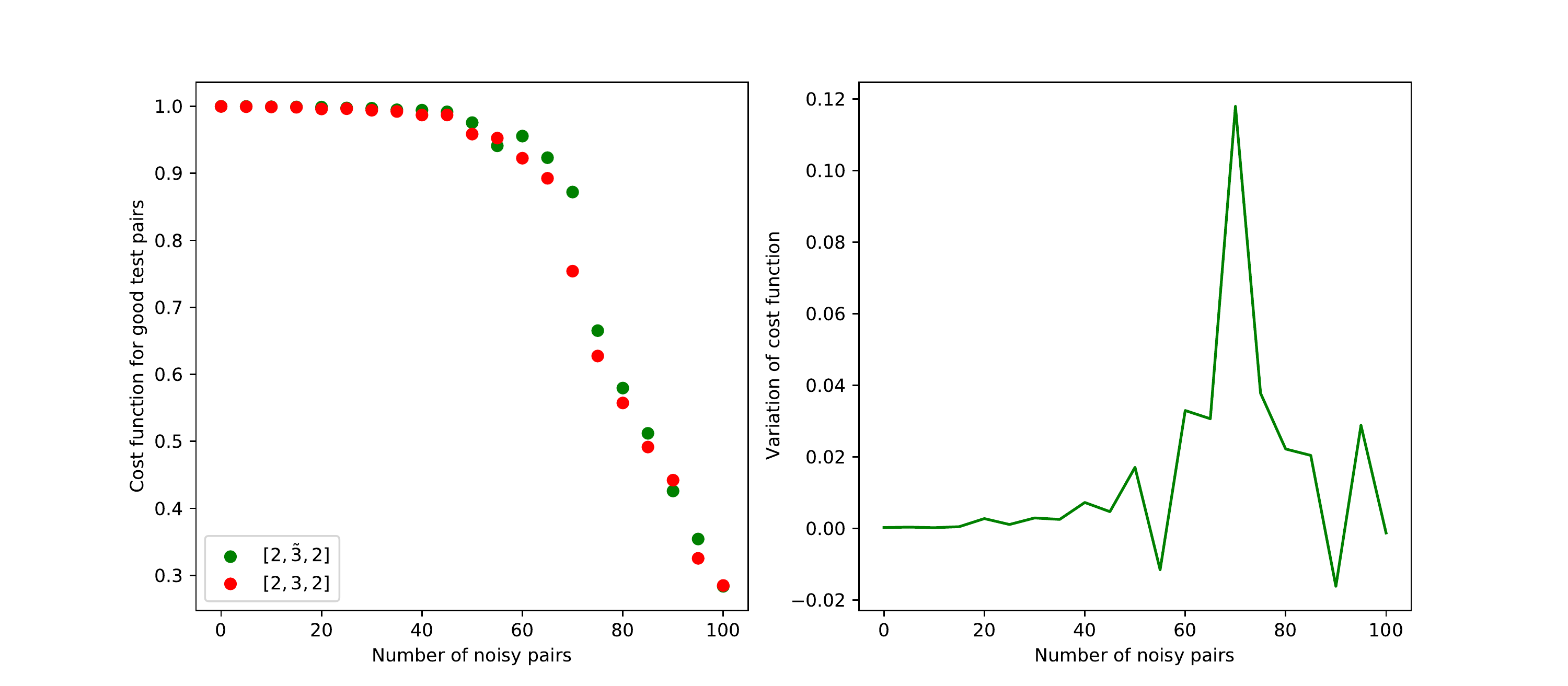}
}
\caption{\textbf{Behaviors of $[2,\tilde{3},2]$ and $[2,3,2]$ to noisy training data.} Here $\eta=1/1.8$ and $\epsilon=0.1$, which is the same setting in Fig. \ref{Figure7}(a). The step-sizes between two adjacent dots in Fig. \ref{Figure11}(a) and Fig. \ref{Figure11}(b) are 3 and 5, respectively. We also plot the variance of cost function between the green dots and red dots on the right. }
\label{Figure11}
\end{figure*}

We choose Res-HQCNN $[2,\tilde{3},2]$ with $\eta=1/1.8$ and $\epsilon=0.1$ as an example and their behaviours are presented in Fig. \ref{Figure11}.
In the left sub-figures of Fig. \ref{Figure11}, the green dots are the results of Res-HQCNN $[2,\tilde{3},2]$ and the red ones are the corresponding results in \cite{beer2020training}. And in the right sub-figures of Fig. \ref{Figure11}, we also plot the variation of the cost function between the green dots and the red ones. The x-axis of Fig. \ref{Figure11} represents how many good training pairs are replaced by noisy pairs.

When the number of the training rounds and training pairs are small, such as 50 training rounds and 30 training pairs in Fig. \ref{Figure11}(a), the values of cost for $[2,\tilde{3},2]$ and $[2,3,2]$ both decrease as the number of noisy pairs increase and the variation of the cost value is always positive. This shows the superiority of $[2,\tilde{3},2]$ for noisy training data with small training rounds and small training pairs.

When the number of training rounds and training pairs are big, such as 200 training rounds and 100 training pairs in Fig. \ref{Figure11}(b), we find that if the number of noisy pairs are small, such as less than 35, Res-HQCNN and QNNs in \cite{beer2020training} both have strong robustness to noisy quantum data. As the number of noisy pairs increasing continuously, the values of cost for green dots and red dots become decrease.

When the number of noisy pairs exceeding 60, the variation of cost becomes increasing, reaching maximum value when the number of noisy pairs are $70$. There are three unstable points $(55,-0.0115)$, $(90,-0.0161)$ and $(100,-0.0012)$ that the variation is negative.
As we can see, there are 21 pairs of green dots and red dots in Fig. \ref{Figure11}. Then the training data are generated  21 times. For each time, the good training data $(\ket{\phi_x^{in}},V\ket{\phi_x^{in}})$ and noisy training data $(\ket{\phi_x^{in}},\ket{\theta_x^{out}})$ are generated randomly. The elements of $\ket{\phi_x^{in}}$ and $\ket{\theta_x^{out}}$ are randomly picked out of a normal distribution before normalization. So we think the randomness of training data causes some unstable points, which can be seen as comparable results. As a whole, Res-HQCNN $[2,\tilde{3},2]$ shows its stronger robustness to noisy data than QNNs $[2,3,2]$.

\begin{figure*}[ht]
\centering
\subfigure[30 training pairs and 150 training rounds]{
\includegraphics[angle=0,width=0.7\linewidth]{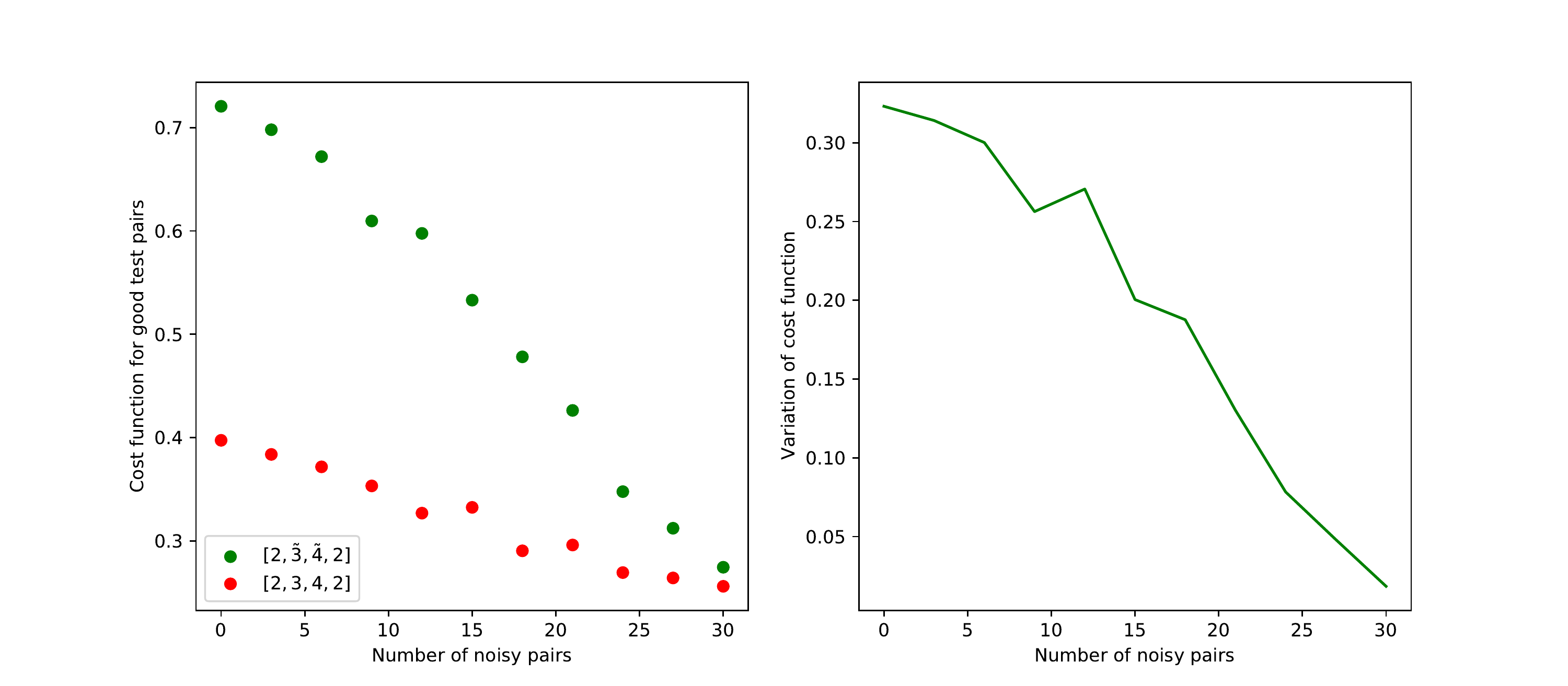}
}
\subfigure[100 training pairs and 600 training rounds]{
\includegraphics[angle=0,width=0.7\linewidth]{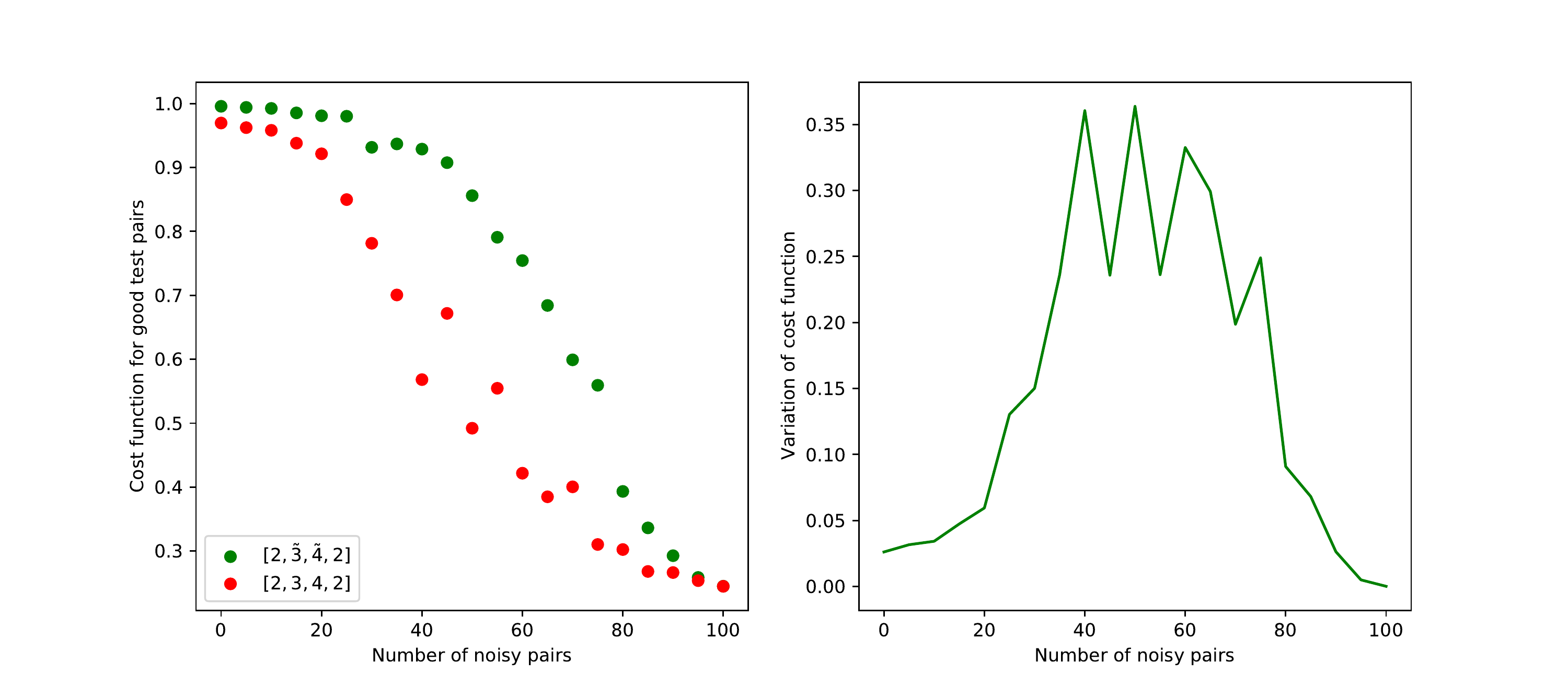}
}
\caption{\textbf{Behaviors of $[2,\tilde{3},\tilde{4}, 2]$ and $[2,3,4,2]$ to noisy training data.} Here $\eta=1/9$ and $\epsilon=0.1$, which is the same setting in Fig. \ref{Figure9}(a). The step-sizes between two adjacent dots are 3 and 5, respectively. We also plot the variance of cost between the green dots and red dots on the right. }
\label{Figure12}
\end{figure*}

We also detect deeper network $[2,\tilde{3},\tilde{4}, 2]$ to noisy data, see Fig. \ref{Figure12}. When the number of training rounds and training pairs are small, such as 150 training rounds and 30 training pairs in Fig. \ref{Figure12}(a), we find that the improvement is obvious. The variances of cost function are always positive. Roughly speaking, the variation decreases with the number of noisy pairs increasing. When the number of training rounds and training pairs are big, such as 600 training rounds and 100 training pairs in Fig. \ref{Figure12}(b), it is pleased to find that all variances of cost function are always positive without unstable points. The maximum value of variance is more than 0.35, while the one in Fig. \ref{Figure11}(b) is no more than 0.12. So for noisy data, deeper $[2,\tilde{3},\tilde{4}, 2]$ shows better improvement than $[2,\tilde{3}, 2]$. 

Up to now, we have gone through the experiments for Res-HQCNN with or without noisy and obtained the improvement of performance for cost function compared with the QNNs in \cite{beer2020training}. Although we do not show the results for Res-HQCNN with four or more hidden layers, we conjecture that deeper Res-HQCNN would bring better improvement of performance for cost function due to the mechanism of its training algorithm.  

\section{Conclusion and discussion}\label{6}
In this paper, we have developed a hybrid quantum-classical neural network with deep residual learning to improve the performance of cost function for deeper networks. A new residual block structure in quantum concept has been designed based on QNNs in \cite{beer2020training}. We have presented how to connect residual block structure with QNNs. The corresponding training algorithm of Res-HQCNN has also been given for different cases. From the perspective of propagating information, the residual block structure allows information propagated from input layer to any deeper layer, which is similar to the mechanism of ANNs with deep residual learning. The simulations have illustrated the power of Res-HQCNN at the cost of only running on classical computer.

\begin{figure*}[ht]
\centering
\subfigure[$p=0.3$]{
\includegraphics[width=6cm]{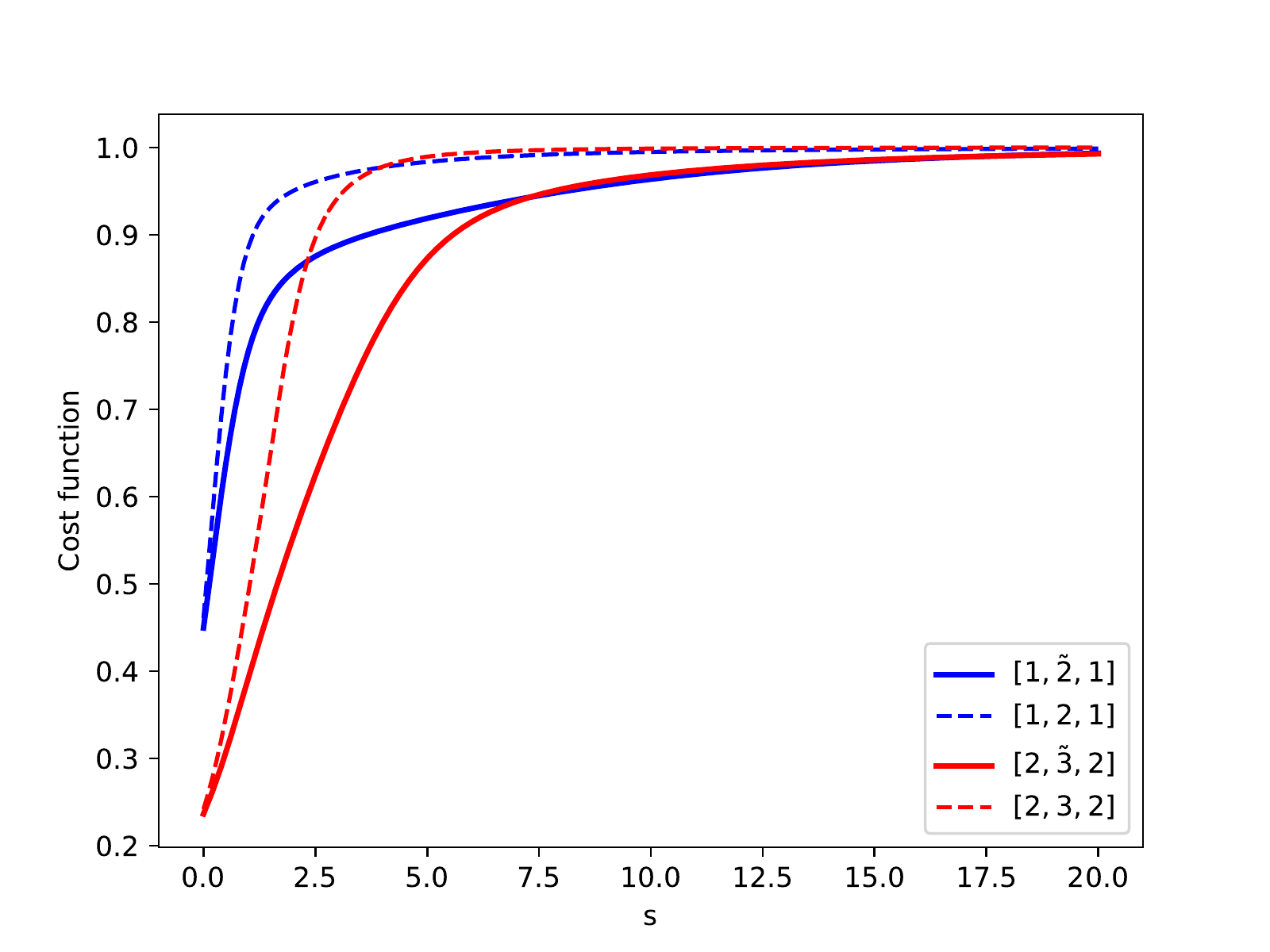}
}
\subfigure[$p=0.6$]{
\includegraphics[width=6cm]{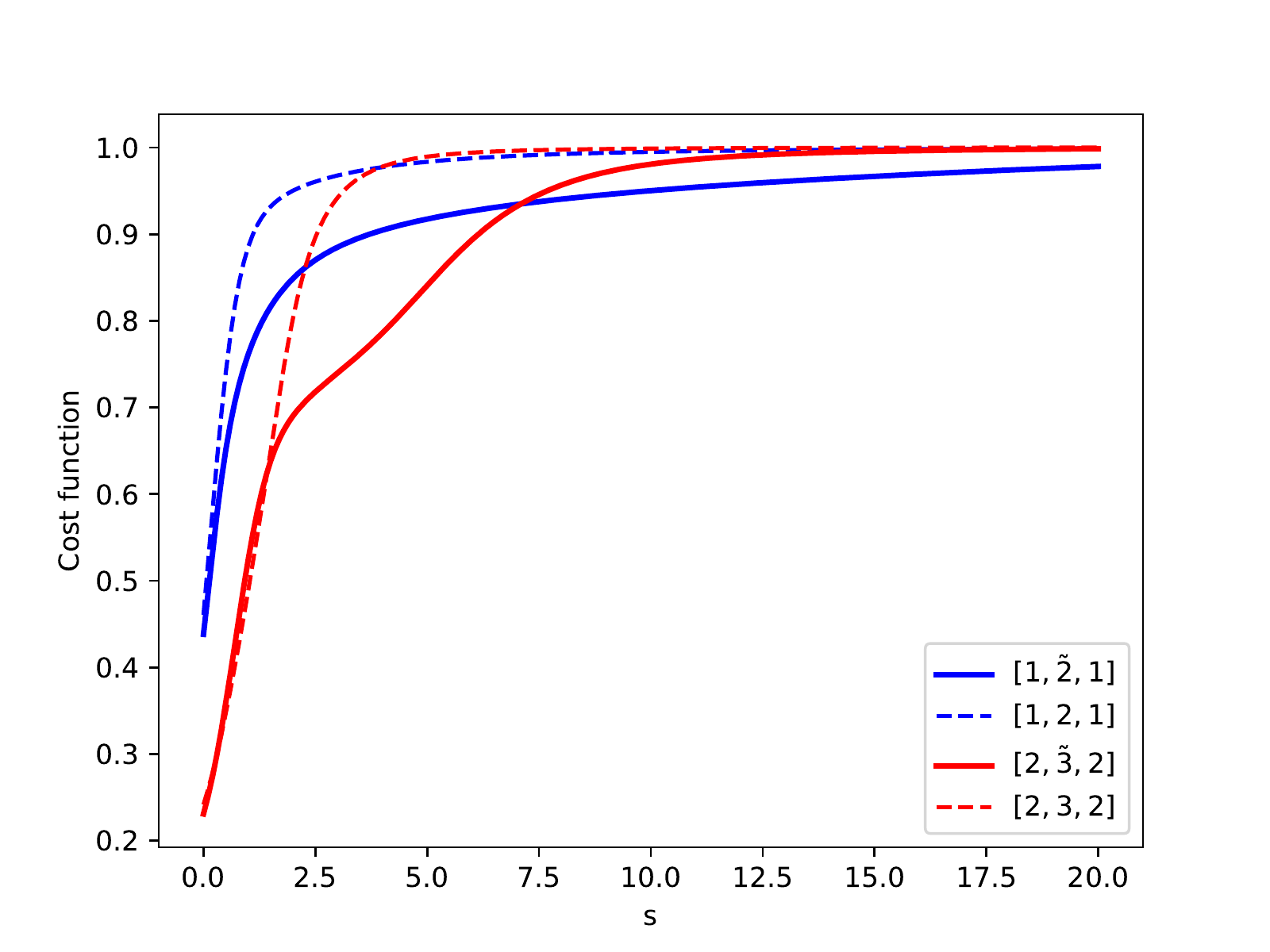}
}
\subfigure[$p=0.9$]{
\includegraphics[width=6cm]{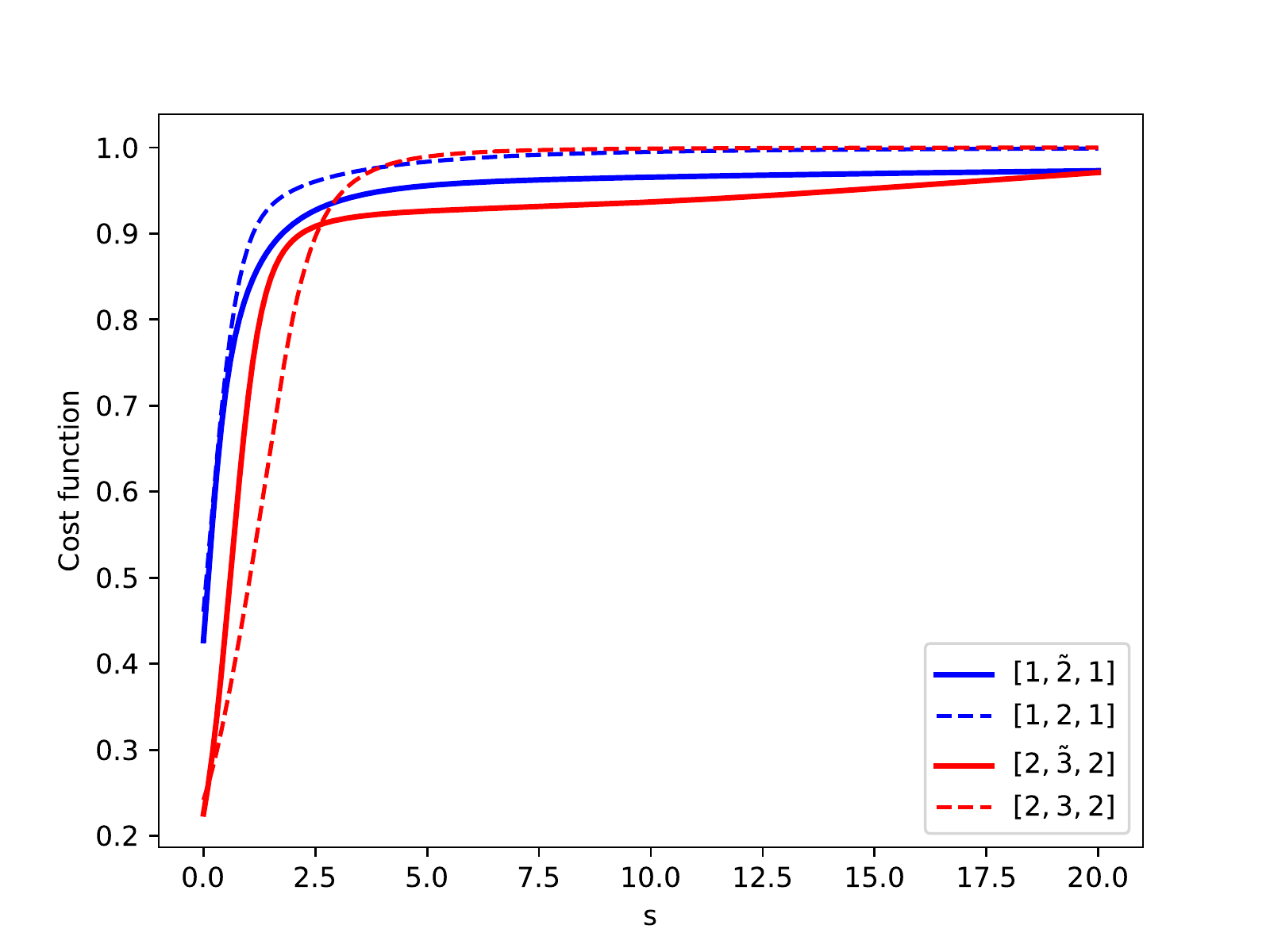}
}
\subfigure[$p=1$]{
\includegraphics[width=6cm]{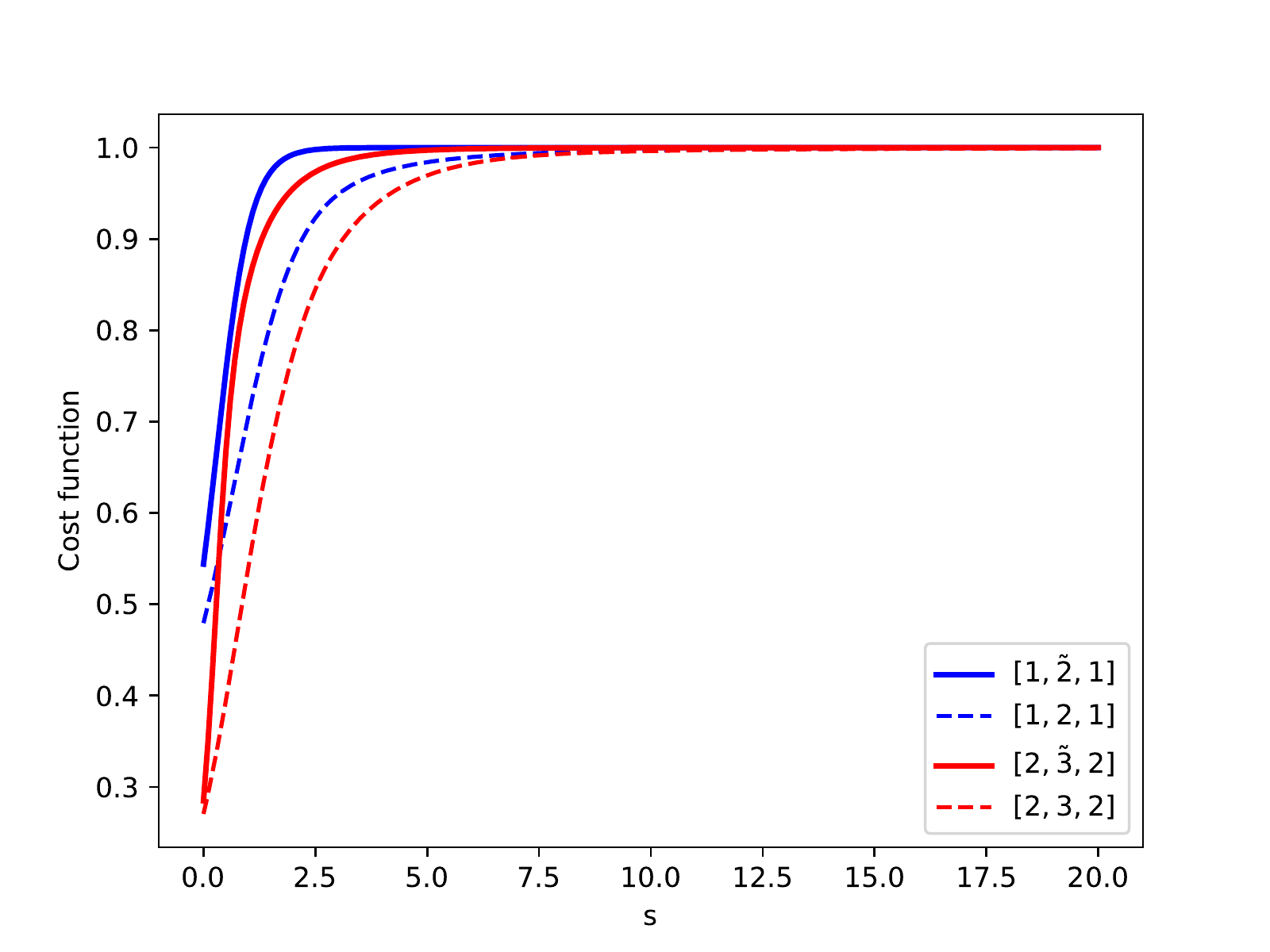}
}
\caption{\textbf{Behaviors of $[1,\tilde{2},1]$ and $[2,\tilde{3},2]$ with different $p$.} Here $\lambda=1$ and $\epsilon=0.1$.  }
\label{Figure13}
\end{figure*}

There is an another method to design residual block structure in quantum concept:
$$\rho^{{l+1}_{in}}=p*\left(\rho^{l_{in}}\otimes\ket{0\cdots0}_{\Delta m_l}\bra{0\cdots0}\right)+(1-p)*\rho^{l_{out}}.$$ It is a convex combination of $\rho^{l_{in}}$ and $\rho^{l_{out}}$.
$\Delta m_l=m_l-m_{l-1}.$
Such an operation is possible to implement on a quantum computer. A quantum device would choose the one state with some probability $p$ and the other with $1-p$ for $0\leq p\leq 1$.
For convenient, we call quantum neural networks with this kind of deep residual learning as p-ResQNN.
The training algorithm of p-ResQNN is also changed. For example, if p-ResQNN has one hidden layer, when $l=1$, the update parameters matrix $K_{j}^{l}(s)=\eta \frac{2^{m_{l-1}}}{N}\sum_{x=1}^{N}\trace_{rest}(1-p)*M_{j}^{l}$; when $l=2$, $K_{j}^{l}(s)=\eta \frac{2^{m_{l-1}}}{N}\sum_{x=1}^{N}\trace_{rest}\left((1-p)*M_{j}^{l}+p*N_{j}^{l}\right)$. $M_{j}^{l}$ and $N_{j}^{l}$ are in Eq. (\ref{k1}) and Eq. (\ref{K2}). The cost function of  p-ResQNN is $C(s)=\frac{1}{N}\sum_{x=1}^{N}\bra{\phi_{x}^{out}}\rho_{x}^{out}(s)\ket{\phi_{x}^{out}},$ which is the same as the definition of cost function in \cite{beer2020training}.
Here are the simulation results of p-ResQNN with one hidden layer, see Fig. \ref{Figure13}.

We choose QNNs $[1,2,1]$ and $[2,3,2]$ to have a test. When $p=0$, it is just the case of former QNNs. 
Randomly choose $p=0.3$, $p=0.6$ $p=0.9$ and $p=1$. From Fig. \ref{Figure13}, we find that
only the result of $p=1$, i.e. $\rho^{{l+1}_{in}}=\rho^{{l}_{in}}$ for $l=1,2,\cdots,L$, can bring improvement of cost function. So we give up this method and do not try to satisfy the requirements for running on a quantum computer. We think
improvement of the performance for cost function are more meaningful for us. 

This paper is mainly focus on how to combine deep residual learning with QNNs to improve the ability of QNNs learning an unknown unitary transform with or without noisy. The readers can learn the advantages and disadvantages from our analysis of the model Res-HQCNN. This is just the beginning of combining quantum neural network with residual block structure. Related work is none. We hope our paper could be an useful reference in this area, which is exactly the point of this paper.

One future investigation of this field could be achieved by processing classical data, such as images and sounds. A suitable data encoding rule between quantum states and real world data should be designed. At present, scholars have carried out many methods on quantum image representation 
\cite{review1image,review2image,review3image}. Due to the combination of quantum computing and deep residual learning, we believe the exploration of Res-HQCNN with real world data will be interesting on some computer vision tasks. Another possible future research is to connect and compare the current results with quantum process tomography \cite{QPT1,QPT2}. Since the main goal in the whole filed of quantum process tomography is to reconstruct quantum processes. Finding out the connection and difference between Res-HQCNN and quantum process tomography may be helpful in characterization of quantum dynamical systems. 

\section*{Declaration of competing interest}
The authors declare that they have no known competing financial interests or personal relationships that could have appeared to influence the work reported in this paper.

\section*{Acknowledgements}
This work is supported by the Guangdong Basic and Applied Basic Research Foundation under Grant No. 2020B1515310016, the Key Research and Development Project of Guangdong Province under Grant No. 2020B0303300001, the Academy of Finland for ICT 2023 project (grant 328115) and project MiGA (grant 316765) and Infotech Oulu. This work is also supported by the Chinese Scholarship Council.

\bibliographystyle{elsarticle-harv}
\bibliography{000amain}

\begin{appendices}
Here we present the training algorithm of a Res-HQCNN with one hidden layer in Fig.\ref{fig5}, which can be denoted as $[2,\tilde{3},2]$. In this training algorithm, we also show the method to calculate the update parameters matrix $K_j^1(s)$ in detail.

\begin{enumerate}
\item[$\mathbf{I}$.] Initialize:
\begin{enumerate}
\item[$\mathbf{I1.}$]Set step $s=0$.
\item[$\mathbf{I2.}$]Choose all unitary $U_{j}^{1}(0)$ and $U_{q}^{1}(0)$ randomly, $j=1,2,3$, $q=1,2$.
\end{enumerate}

\item[$\mathbf{II}$.] For each element $(\ket{\phi_{x}^{in}}, \ket{\phi_{x}^{out}})$ from the set of training data, do the following steps:
\begin{enumerate}
\item[$\mathbf{II1.}$]Feedforward:
\begin{enumerate}
\item[$\mathbf{II1a.}$] Tensor the input state $\rho_x^{1_{in}}=\ket{\phi_{x}^{in}}\bra{\phi_{x}^{in}}$ to the initial state of the hidden layer,
$$\ket{\phi_{x}^{in}}\bra{\phi_{x}^{in}}\otimes\ket{000}_{hid}
\bra{000}.$$
\item[$\mathbf{II1b.}$] Apply the layer unitary $U^{1}(s)=U_3^{1}U_2^{1}U_{1}^{1}(s)$,
$$U_{apply}^{1}(s)=U^{1}(s)\left(\ket{\phi_{x}^{in}}\bra{\phi_{x}^{in}}\otimes\ket{000}_{hid}\bra{000}\right){U^{1}}^{\dagger}(s).$$
\item[$\mathbf{II1c.}$] Trace out the input layer and obtain the output state of the hidden layer,
$$\rho_x^{1_{out}}(s)=\trace_{in}(U_{apply}^{1}(s)).$$
\end{enumerate}

\item[$\mathbf{II.2}$]Residual learning:
\begin{enumerate}
\item[$\mathbf{II2a.}$] Apply the residual block structure to $\rho_x^{1_{out}}(s)$  to get the new input state of layer with $l=2$, namely the final output layer,
$$\rho_x^{2_{in}}(s)=\rho_x^{1_{in}}(s)\otimes\ket{0}\bra{0}+\rho_x^{1_{out}}(s).$$
\item[$\mathbf{II2b.}$]Store $\rho_x^{2_{in}}(s)$.
\end{enumerate}
\item[$\mathbf{II3.}$]Repeat Step II.1 for  $\rho_x^{2_{in}}(s)$, and get the output of layer with $l=2$, 
$$\rho_x^{2_{out}}(s)=\trace_{hid}(U^{2}(s)\left(\rho_x^{2_{in}}(s)\otimes \ket{00}_{out}\bra{00}\right){U^{2}}^{\dagger}(s)).$$

\end{enumerate}
\item[$\mathbf{III}$.] Update the parameters:
\begin{enumerate}
\item[$\mathbf{III1.}$]Compute the cost function:
$$C(s)=\frac{1}{2N}\sum_{x=1}^{N}\bra{\phi_{x}^{out}}\rho_x^{out}(s)\ket{\phi_{x}^{out}}.$$
\item[$\mathbf{III2.}$]Calculate the parameter matrices $K_j^1(s)$ and $K_q^2(s)$ for $j=1,2,3$ and $q=1,2$, which will be illustrated later.
\item[$\mathbf{III3.}$]Update each perceptron unitary via
$$U_{j}^{1}(s+\epsilon)=e^{i\epsilon K_{j}^{1}(s)}U_{j}^{1}(s). $$
$$U_{q}^{2}(s+\epsilon)=e^{i\epsilon K_{q}^{2}(s)}U_{q}^{2}(s). $$
\item[$\mathbf{III4.}$]Update $s=s+\epsilon$.
\end{enumerate}

\item[$\mathbf{IV}$.] Repeat Step II and Step III until reaching the maximum of the cost function.
\end{enumerate}
In the following, we will derive a formula for $K_{j}^{l}(s)$ to update the perceptron unitaries $U_{j}^{l}(s)$ with $l=1,2$.  We assume the unitaries can always act on its current layer, such as $U_{2}^{1}(s)$ is actually $U_{2}^{1}(s)\otimes \ket{00}\bra{00}$ in Res-HQCNN $[2,\tilde{3},2]$. Inspired by the method from \cite{beer2020training}, we present a different way to calculate $K_{j}^{l}(s)$ with the residual block structure.

We begin with a mathematical definition of derivative function for the cost function: $\frac{dC}{ds}=\lim\limits_{\epsilon \to 0}{\frac{C(s+\epsilon)-C(s)}{\epsilon}}$,
which leads us to find an analytical expression of $C(s+\epsilon)$. According to the definition of $C(s+\epsilon)$, we focus on the output state of the updated unitary
\begin{align}\label{a1}
\setcounter{equation}{0}
\renewcommand\theequation{A.\arabic{equation}}
U^{1}(s+\epsilon)=e^{i\epsilon K_{3}^{1}(s)}U_{3}^{1}(s)e^{i\epsilon K_{2}^{1}(s)}U_{2}^{1}(s)e^{i\epsilon K_{1}^{1}(s)}U_{1}^{1}(s), 
\end{align}
\begin{align}\label{a2}
U^{2}(s+\epsilon)=e^{i\epsilon K_{2}^{2}(s)}U_{2}^{2}(s)e^{i\epsilon K_{1}^{2}(s)}U_{1}^{2}(s).
\end{align}
Firstly, we consider the output state of the updated unitary $U^{1}(s+\epsilon)$ in the hidden layer. For convenience, we omit to write the parameter $s$ in $K_{j}^{1}(s)$ and $U_{j}^{1}(s)$ in the following with $j=1,2,3$. Then using Eq.(\ref{a1}), we have
\begin{align}\label{a3}
    \rho_x^{1_{out}}(s+\epsilon)=&\trace_{in}(e^{i\epsilon K_{3}^{1}}U_{3}^{1} e^{i\epsilon K_{2}^{1}}U_{2}^{1} e^{i\epsilon K_{1}^{1}}U_{1}^{1} \left(\ket{\phi_{x}^{in}}\bra{\phi_{x}^{in}}\otimes\ket{000}_{hid}\bra{000}\right){U_{1}^{1}}^{\dagger}e^{-i\epsilon K_{1}^{1}}\nonumber\\
    &{U_{2}^{1}}^{\dagger}e^{-i\epsilon K_{2}^{1}} {U_{3}^{1}}^{\dagger}e^{-i\epsilon K_{3}^{1}})\nonumber\\
    =&\rho_x^{1_{out}}(s)+i\epsilon \trace_{in}(U_{3}^{1}U_{2}^{1}K_{1}^{1}U_{1}^{1}\left(\ket{\phi_{x}^{in}}\bra{\phi_{x}^{in}}\otimes\ket{000}_{hid}\bra{000}\right){U_{1}^{1}}^{\dagger}{U_{2}^{1}}^{\dagger}{U_{3}^{1}}^{\dagger}\nonumber\\
    &+U_{3}^{1}K_{2}^{1}U_{2}^{1}U_{1}^{1}\left(\ket{\phi_{x}^{in}}\bra{\phi_{x}^{in}}\otimes\ket{000}_{hid}\bra{000}\right){U_{1}^{1}}^{\dagger}{U_{2}^{1}}^{\dagger}{U_{3}^{1}}^{\dagger}\nonumber\\
    &+K_{3}^{1}U_{3}^{1}U_{2}^{1}U_{1}^{1}\left(\ket{\phi_{x}^{in}}\bra{\phi_{x}^{in}}\otimes\ket{000}_{hid}\bra{000}\right){U_{1}^{1}}^{\dagger}{U_{2}^{1}}^{\dagger}{U_{3}^{1}}^{\dagger}\nonumber\\
    &
    -U_{3}^{1}U_{2}^{1}U_{1}^{1}\left(\ket{\phi_{x}^{in}}\bra{\phi_{x}^{in}}\otimes\ket{000}_{hid}\bra{000}\right){U_{1}^{1}}^{\dagger}{U_{2}^{1}}^{\dagger}{U_{3}^{1}}^{\dagger}K_{3}^{1}\nonumber\\
    &-U_{3}^{1}U_{2}^{1}U_{1}^{1}\left(\ket{\phi_{x}^{in}}\bra{\phi_{x}^{in}}\otimes\ket{000}_{hid}\bra{000}\right){U_{1}^{1}}^{\dagger}{U_{2}^{1}}^{\dagger}K_{2}^{1}{U_{3}^{1}}^{\dagger}\nonumber\\
    &-U_{3}^{1}U_{2}^{1}U_{1}^{1}\left(\ket{\phi_{x}^{in}}\bra{\phi_{x}^{in}}\otimes\ket{000}_{hid}\bra{000}\right) {U_{1}^{1}}^{\dagger}K_{1}^{1}{U_{2}^{1}}^{\dagger}{U_{3}^{1}}^{\dagger})+o(\epsilon^2)\nonumber\\
    =&\rho_x^{1_{out}}(s)+R(\epsilon)+o(\epsilon^2),
\end{align}
where the second inequality is due to the Taylor's Formula of the exponential function. We denote the second term in the second inequality as $R(\epsilon)$.
Then using the residual block structure to $\rho_x^{1_{out}}(s+\epsilon)$, we obtain the new input state of layer with $l=2$:
\begin{align}\label{a4}
 \rho_x^{2_{in}}(s+\epsilon)=& \rho_x^{1_{out}}(s+\epsilon)+\ket{\phi_x^{in}}\bra{\phi_x^{in}}\otimes \ket{0}\bra{0}\nonumber\\
=&(\rho_x^{1_{out}}(s)+\ket{\phi_x^{in}}\bra{\phi_x^{in}}\otimes \ket{0}\bra{0})+R(\epsilon)+o(\epsilon^2)\nonumber\\
=&\rho_x^{2_{in}}(s)+R(\epsilon)+o(\epsilon^2).
\end{align}

Based on the new input state for layer with $l=2$, we then go on computing the updated final output state of Res-HQCNN $[2,\tilde{3},2]$ in terms of $U^2(s+\epsilon)$ in Eq.(\ref{a2}),
\begin{align}\label{a5}
    \rho_x^{2_{out}}(s+\epsilon)=&\trace_{hid}(e^{i\epsilon K_{2}^{2}}U_{2}^{2} e^{i\epsilon K_{1}^{2}}U_{1}^{2}  \left(\rho_x^{2_{in}}(s+\epsilon)\otimes\ket{00}_{out}\bra{00}\right)
    {U_{1}^{2}}^{\dagger}e^{-i\epsilon K_{1}^{2}} {U_{2}^{2}}^{\dagger}e^{-i\epsilon K_{2}^{2}} )\nonumber\\
    =&\trace_{hid}(e^{i\epsilon K_{2}^{2}}U_{2}^{2} e^{i\epsilon K_{1}^{2}}U_{1}^{2}  \left(\rho_x^{2_{in}}(s)\otimes\ket{00}_{out}\bra{00}\right)
    {U_{1}^{2}}^{\dagger}e^{-i\epsilon K_{1}^{2}} {U_{2}^{2}}^{\dagger}e^{-i\epsilon K_{2}^{2}} )\nonumber\\
    &+\trace_{hid}(e^{i\epsilon K_{2}^{2}}U_{2}^{2} e^{i\epsilon K_{1}^{2}}U_{1}^{2}  \left(R(\epsilon)\otimes\ket{00}_{out}\bra{00}\right){U_{1}^{2}}^{\dagger}e^{-i\epsilon K_{1}^{2}}{U_{2}^{2}}^{\dagger}e^{-i\epsilon K_{2}^{2}})+o(\epsilon^2)\nonumber\\
=&(\rho_x^{2_{out}}(s)+A_1)+A_2+o(\epsilon^2),
\end{align}
in which
\begin{align*}
A_1=&i\epsilon\trace_{hid}(U_2^{2}K_1^{2}U_1^{2}(\rho_x^{2_{in}}(s)\otimes\ket{00}_{out}\bra{00}){U_{1}^{2}}^{\dagger}{U_{2}^{2}}^{\dagger}\nonumber\\
&+K_2^{2}U_2^{2}U_1^{2}(\rho_x^{2_{in}}(s)\otimes\ket{00}_{out}\bra{00}){U_{1}^{2}}^{\dagger}{U_{2}^{2}}^{\dagger}\nonumber\\
&-U_2^{2}U_1^{2}(\rho_x^{2_{in}}(s)\otimes\ket{00}_{out}\bra{00}){U_{1}^{2}}^{\dagger}{U_{2}^{2}}^{\dagger}K_2^{2}\nonumber\\
&-U_2^{2}U_1^{2}(\rho_x^{2_{in}}(s)\otimes\ket{00}_{out}\bra{00}){U_{1}^{2}}^{\dagger}K_1^{2}{U_{2}^{2}}^{\dagger}),
\end{align*}
\begin{align*}
A_2=&i\epsilon\trace_{in,hid}(U_2^{2}U_1^{2}U_{3}^{1}U_{2}^{1}K_{1}^{1}U_{1}^{1}(\ket{\phi_x^{in}}\bra{\phi_x^{in}}\otimes\nonumber\\
&\ket{00000}_{hid,out}\bra{00000}){U_{1}^{1}}^{\dagger}{U_{2}^{1}}^{\dagger}{U_{3}^{1}}^{\dagger}{U_{1}^{2}}^{\dagger}{U_{2}^{2}}^{\dagger}\nonumber\\
&+U_2^{2}U_1^{2}U_{3}^{1}K_{2}^{1}U_{2}^{1}U_{1}^{1}(\ket{\phi_x^{in}}\bra{\phi_x^{in}}\otimes\nonumber\\
&\ket{00000}_{hid,out}\bra{00000}){U_{1}^{1}}^{\dagger}{U_{2}^{1}}^{\dagger}{U_{3}^{1}}^{\dagger}{U_{1}^{2}}^{\dagger}{U_{2}^{2}}^{\dagger}\nonumber\\
&+U_2^{2}U_1^{2}K_{3}^{1}U_{3}^{1}U_{2}^{1}U_{1}^{1}(\ket{\phi_x^{in}}\bra{\phi_x^{in}}\otimes\nonumber\\
&\ket{00000}_{hid,out}\bra{00000}){U_{1}^{1}}^{\dagger}{U_{2}^{1}}^{\dagger}{U_{3}^{1}}^{\dagger}{U_{1}^{2}}^{\dagger}{U_{2}^{2}}^{\dagger}\nonumber\\
&-U_2^{2}U_1^{2}U_{3}^{1}U_{2}^{1}U_{1}^{1}(\ket{\phi_x^{in}}\bra{\phi_x^{in}}\otimes\nonumber\\
&\ket{00000}_{hid,out}\bra{00000}){U_{1}^{1}}^{\dagger}{U_{2}^{1}}^{\dagger}{U_{3}^{1}}^{\dagger}K_{3}^{1}{U_{1}^{2}}^{\dagger}{U_{2}^{2}}^{\dagger})\nonumber\\
&-U_2^{2}U_1^{2}U_{3}^{1}U_{2}^{1}U_{1}^{1}(\ket{\phi_x^{in}}\bra{\phi_x^{in}}\otimes\nonumber\\
&\ket{00000}_{hid,out}\bra{00000}){U_{1}^{1}}^{\dagger}{U_{2}^{1}}^{\dagger}K_{2}^{1}{U_{3}^{1}}^{\dagger}{U_{1}^{2}}^{\dagger}{U_{2}^{2}}^{\dagger})\nonumber\\
&-U_2^{2}U_1^{2}U_{3}^{1}U_{2}^{1}U_{1}^{1}(\ket{\phi_x^{in}}\bra{\phi_x^{in}}\otimes\nonumber\\
&\ket{00000}_{hid,out}\bra{00000}){U_{1}^{1}}^{\dagger}K_{1}^{1}{U_{2}^{1}}^{\dagger}{U_{3}^{1}}^{\dagger}{U_{1}^{2}}^{\dagger}{U_{2}^{2}}^{\dagger}).
\end{align*}
So the mathematical derivative function of the cost function can be calculated as
\begin{align}\label{a6}
\frac{dC}{ds}=&\lim_{\epsilon \to 0}{\frac{C(s+\epsilon)-C(s)}{\epsilon}}\nonumber\\
=&\lim_{\epsilon \to 0}\frac{\frac{1}{N}\sum_{x=1}^{N}\bra{\phi_{x}^{out}}\rho_{x}^{2_{out}}(s+\epsilon)\ket{\phi_{x}^{out}}}{\epsilon}\nonumber\\
&-\lim_{\epsilon \to 0}\frac{\frac{1}{N}\sum_{x=1}^{N}\bra{\phi_{x}^{out}}\rho_{x}^{2_{out}}(s)\ket{\phi_{x}^{out}}}{\epsilon}\nonumber\\
=&\frac{1}{N}\sum_{x=1}^{N}\trace(Id(3)\otimes\ket{\phi_{x}^{out}}\bra{\phi_{x}^{out}}([i K_2^2,U_2^2U_1^2(\rho_x^{2_{in}}\otimes\ket{00}_{out}\bra{00}{U_1^2}^{\dagger}{U_2^2}^{\dagger}]+U_2^2\nonumber\\
&[i K_1^2,U_1^2(\rho_x^{2_{in}}\otimes\ket{00}_{out}\bra{00}{U_1^2}^{\dagger}]{U_2^2}^{\dagger}+\cdots+U_2^1U_3^1U_1^2U_2^2[i K_1^1, U_1^1(\rho_x^{1_{in}}\otimes\ket{000}_{hid}\bra{000})\nonumber\\
&{U_1^1}^{\dagger}]{U_2^2}^{\dagger}{U_1^2}^{\dagger}{U_3^1}^{\dagger}{U_2^1}^{\dagger}
))\nonumber\\
=&\frac{i}{N}\sum_{x=1}^{N}\trace(([U_2^2U_1^2(\rho_x^{2_{in}}\otimes\ket{00}_{out}\bra{00}{U_1^2}^{\dagger}{U_2^2}^{\dagger},Id(3)\otimes\ket{\phi_{x}^{out}}\bra{\phi_{x}^{out}}]K_2^2+\nonumber\\
&[U_1^2(\rho_x^{2_{in}}\otimes\ket{00}_{out}\bra{00}{U_1^2}^{\dagger},{U_2^2}^{\dagger}(Id(3)\otimes\ket{\phi_{x}^{out}}\bra{\phi_{x}^{out}})U_2^2]K_1^2+\cdots+\nonumber\\
&[ U_1^1(\rho_x^{1_{in}}\otimes\ket{000}_{hid}\bra{000}){U_1^1}^{\dagger},{U_2^1}^{\dagger}{U_3^1}^{\dagger}{U_1^2}^{\dagger}{U_2^2}^{\dagger}(Id(3)\otimes\ket{\phi_{x}^{out}}\bra{\phi_{x}^{out}})U_2^2U_1^2U_3^1U_2^1]K_1^1)\nonumber\\
=&\frac{i}{N}\sum_{x=1}^{N}\trace\left((M_2^2+N_2^2)K_2^2+(M_1^2+N_1^2)K_1^2\right)+\nonumber\\
&\frac{i}{N}\sum_{x=1}^{N}\trace(M_3^1K_3^1+M_2^1K_2^1+M_1^1K_1^1),
\end{align}
where 
\begin{align*}
M_1^1=&[U_1^1((\ket{\phi_x^{in}}\bra{\phi_x^{in}}\otimes\ket{000}_{hid}\bra{000}){U_1^1}^{\dagger},\nonumber\\
&{U_2^1}^{\dagger}{U_3^1}^{\dagger}{U_1^2}^{\dagger}{U_2^2}^{\dagger}(Id(3)\otimes\ket{\phi_{x}^{out}}\bra{\phi_{x}^{out}})U_2^2U_1^2U_2^1U_3^1],
\end{align*}
\begin{align*}
M_2^1=&[U_2^1U_1^1((\ket{\phi_x^{in}}\bra{\phi_x^{in}}\otimes\ket{000}_{hid}\bra{000}){U_1^1}^{\dagger}{U_2^1}^{\dagger},\nonumber\\
&{U_3^1}^{\dagger}{U_1^2}^{\dagger}{U_2^2}^{\dagger}(Id(3)\otimes\ket{\phi_{x}^{out}}\bra{\phi_{x}^{out}})U_2^2U_1^2U_3^1],
\end{align*}
\begin{align*}
M_3^1=&[U_3^1U_2^1U_1^1((\ket{\phi_x^{in}}\bra{\phi_x^{in}}\otimes\ket{000}_{hid}\bra{000}){U_1^1}^{\dagger}{U_2^1}^{\dagger}{U_3^1}^{\dagger},\nonumber\\
&{U_1^2}^{\dagger}{U_2^2}^{\dagger}(Id(3)\otimes\ket{\phi_{x}^{out}}\bra{\phi_{x}^{out}})U_2^2U_1^2],
\end{align*}
\begin{align*}
M_1^2=&[U_1^2U_3^1U_2^1U_1^1((\ket{\phi_x^{in}}\bra{\phi_x^{in}}\otimes\ket{000}_{hid}\bra{000}){U_1^1}^{\dagger}{U_2^1}^{\dagger}{U_3^1}^{\dagger}{U_1^2}^{\dagger},\nonumber\\
&{U_2^2}^{\dagger}(Id(3)\otimes\ket{\phi_{x}^{out}}\bra{\phi_{x}^{out}})U_2^2],
\end{align*}
\begin{align*}
M_2^2=&[U_2^2U_1^2U_3^1U_2^1U_1^1((\ket{\phi_x^{in}}\bra{\phi_x^{in}}\otimes\ket{000}_{hid}\bra{000}){U_1^1}^{\dagger}{U_2^1}^{\dagger}{U_3^1}^{\dagger}{U_1^2}^{\dagger}{U_2^2}^{\dagger},\nonumber\\
&Id(3)\otimes\ket{\phi_{x}^{out}}\bra{\phi_{x}^{out}}],
\end{align*}
\begin{align*}
N_1^2=&[U_1^2((\ket{\phi_x^{in}}\bra{\phi_x^{in}}\otimes\ket{000}_{hid}\bra{000}){U_1^2}^{\dagger},
{U_2^2}^{\dagger}(Id(3)\otimes\ket{\phi_{x}^{out}}\bra{\phi_{x}^{out}})U_2^2],
\end{align*}
\begin{align*}
N_2^2=&[U_2^2U_1^2((\ket{\phi_x^{in}}\bra{\phi_x^{in}}\otimes\ket{000}_{hid}\bra{000}){U_1^2}^{\dagger}{U_2^2}^{\dagger},
Id(3)\otimes\ket{\phi_{x}^{out}}\bra{\phi_{x}^{out}}],
\end{align*}
Here the third equality in Eq.(\ref{a6}) is due to Eq.(\ref{a5}) and the property $\bra{\phi}A\ket{\phi}=tr(A\ket{\phi}\bra{\phi})$. The fourth equality is obtained by the property $tr(AB)=tr(BA)$. $A$ and $B$ represent arbitrary matrix. Using Eq.(\ref{a4}), we get the last equality.
The mathematical commutator operator is in unitary group, which reads $[a,b]=a\times b-b\times a$ for arbitrary unitary matrix $a$ and $b$. The ''$\times$`` means the multiplication of matrix. This symbol is omitted above.

Up to now, we get the analytical expression of $\frac{dC}{ds}$, which can be seen as a linear function of $K_j^l$. To find the maximum of the cost function fastest, we should maximize $\frac{dC}{ds}$. When $l=1$, since unitary $U_j^1(s)$ in the quantum perceptron works on three qubits, then we can parameter  $K_j^1$ as
\begin{align}\label{a7}
K_j^1(s)=\sum_{\alpha_1,\alpha_2,\beta}{K_j^1}_{\alpha_1,\alpha_2,\beta}(s)(\sigma^{\alpha_1}\otimes\sigma^{\alpha_2}\otimes\sigma^{\beta}),
\end{align}
where $\sigma$ is Pauli matrix in single qubit. $\alpha_1$ and $\alpha_2$ represent the qubits in the input layer and $\beta$ represents the current qubit of unitary $U_j^1$ in the hidden layer. We choose to use of a Lagrange multiplier $\lambda$, which is a real number, to find a finite solution. When $j=1$, the analysis above leads us to solve a maximization problem in the following:
\begin{equation*}
\begin{split}
&\max_{K_1^1,\alpha_1,\alpha_2,\beta}(\frac{dC}{ds}-\lambda\sum_{\alpha_1,\alpha_2,\beta}{{K_1^1}_{\alpha_1,\alpha_2,\beta}}^2)\\
&=\max_{K_1^1,\alpha_1,\alpha_2,\beta}(\frac{i}{N}\sum_{x=1}^{N}\trace((M_1^2+N_1^2)K_1^2+(M_2^2+N_2^2)K_2^2+M_3^1K_3^1
+M_2^1K_2^1\\
&+M_1^1K_1^1
- \lambda\sum_{\alpha_1,\alpha_2,\beta}{{K_1^1}_{\alpha_1,\alpha_2,\beta}}^2 )\\
&=\max_{K_1^1,\alpha_1,\alpha_2,\beta}(\frac{i}{N}\sum_{x=1}^{N}\trace_{\alpha_1,\alpha_2,\beta}[\trace_{rest}\left((M_1^2+N_1^2)K_1^2+(M_2^2+N_2^2)K_2^2
+M_3^1K_3^1+M_2^1K_2^1\right)\\
&+\trace_{rest}(M_1^1)\sum_{\alpha_1,\alpha_2,\beta}{K_1^1}_{\alpha_1,\alpha_2,\beta}(\sigma^{\alpha_1}
\otimes\sigma^{\alpha_2}\otimes\sigma^{\beta})]
- \lambda\sum_{\alpha_1,\alpha_2,\beta}{{K_1^1}_{\alpha_1,\alpha_2,\beta}}^2 )\\
\end{split}
\end{equation*}
Here $tr_{rest}$ means the complement of $\{\alpha_1,\alpha_2,\beta\}$. The second equality is obtained by Eq.(\ref{a7}).
Taking the derivative of ${K_j^1}_{\alpha_1,\alpha_2,\beta}$ to be zero yields
$${K_1^1}_{\alpha_1,\alpha_2,\beta}=\frac{i}{2\lambda N}\sum_{x=1}^{N}\trace_{\alpha_1,\alpha_2,\beta}\left((\trace_{rest}(M_1^1))(\sigma^{\alpha_1}\otimes\sigma^{\alpha_2}\otimes\sigma^{\beta})\right).$$
This above equation further leads to the matrix:
\begin{align*}
K_1^1=&\sum_{\alpha_1,\alpha_2,\beta}{K_1^1}_{\alpha_1,\alpha_2,\beta}(\sigma^{\alpha_1}\otimes\sigma^{\alpha_2}\otimes\sigma^{\beta})\nonumber\\
=&\frac{i}{2\lambda N}\sum_{x=1}^{N}\sum_{\alpha_1,\alpha_2,\beta}\trace_{\alpha_1,\alpha_2,\beta}\left((\trace_{rest}(M_1^1))(\sigma^{\alpha_1}\otimes\sigma^{\alpha_2}\otimes\sigma^{\beta})
\times(\sigma^{\alpha_1}\otimes\sigma^{\alpha_2}\otimes\sigma^{\beta})\right)\nonumber\\
=&\frac{4i}{\lambda N}\sum_{x=1}^{N}\trace_{rest}(M_1^1).
\end{align*}
Here $\frac{1}{\lambda}$ is regarded as the learning rate in this paper.
Analogously, we find out the formulas for $K_2^1$ and $K_3^1$:
$K_2^1=\frac{4i}{\lambda N}\sum_{x=1}^{N}\trace_{rest}(M_2^1);$
$K_3^1=\frac{4i}{\lambda N}\sum_{x=1}^{N}\trace_{rest}(M_3^1).$

As $K_q^2(s)$ in Res-HQCNN $[2,\tilde{3},2]$ with $q=1,2$ can be parameterized as
$$K_q^2(s)=\sum_{\alpha_1,\alpha_2,\alpha_3,\beta}{K_q^2}_{\alpha_1,\alpha_2,\alpha_3,\beta}(s)(\sigma^{\alpha_1}\otimes\sigma^{\alpha_2}\otimes\sigma^{\alpha_3}\otimes\sigma^{\beta}),$$
in which $\alpha_1,\alpha_2,\alpha_3$ represents the qubits in the hidden layer and $\beta$ represents the current qubit of unitary $U_q^2$ in the output layer.
After going on the similar process for $K_j^1$, we can have the formula for $K_q^2$:
$$K_q^2=\frac{8i}{\lambda N}\sum_{x=1}^{N}\trace_{rest}(M_q^2+N_q^2).$$

\end{appendices}

\end{document}